\documentclass{article} 
\usepackage{arxiv_style,times}

\usepackage{amsmath,amsfonts,bm}

\def\eqref#1{equation~\ref{#1}}

\def\1{\bm{1}}

\DeclareMathAlphabet{\mathsfit}{\encodingdefault}{\sfdefault}{m}{sl}
\SetMathAlphabet{\mathsfit}{bold}{\encodingdefault}{\sfdefault}{bx}{n}

\usepackage{hyperref}
\usepackage{url}
\usepackage{booktabs}
\usepackage{colortbl}
\usepackage{amssymb}
\usepackage{multirow}
\usepackage{graphicx}
\usepackage{subcaption}
\usepackage{placeins}
\definecolor{arxivlinkblue}{RGB}{79,122,159}
\hypersetup{pdfborder={0 0 0}}

\makeatletter
\def\fps@table{!tp}
\makeatother

\title{Looking Ahead: Modeling Future Evolution \\ in Autoregressive Video Generation}

\iclrfinalcopy
\author{Ruopeng Gao$^{1,*}$, Yukun Huang$^{2,\ddagger}$, Jiarui Hu$^{3}$, Shuai Wang$^{1}$,\\
  \textbf{Wenbo Hu$^{2}$, Ying Shan$^{2}$, Limin Wang$^{1,\dagger}$}\\[0.5ex]
  $^{1}$Nanjing University \quad $^{2}$ARC Lab, Tencent IEG \quad $^{3}$Zhejiang University\\[0.5ex]
  {\normalfont\begin{tabular}{@{}l@{\quad}l@{}}
    \textbf{Project page:} &
    \href{https://hellorpg.github.io/looking-ahead/}{\textcolor{arxivlinkblue}{\texttt{https://hellorpg.github.io/looking-ahead/}}}\\[0.2ex]
    \textbf{Code:} &
    \href{https://github.com/TencentARC/Looking-Ahead}{\textcolor{arxivlinkblue}{\texttt{https://github.com/TencentARC/Looking-Ahead}}}
  \end{tabular}}}
\makeatletter
\gdef\@thanks{%
  \footnotetext[1]{Work done during internship at ARC Lab, Tencent IEG.}%
  \footnotetext[2]{Corresponding author.}%
  \footnotetext[3]{Project lead.}%
}
\makeatother

\begin{document}

\maketitle

\begin{abstract}
Autoregressive video diffusion models generate videos chunk by chunk, yet producing plausible continuations does not guarantee that they capture how scenes evolve.
Under teacher forcing, strong temporal redundancy may allow local visual continuity to serve as a shortcut for current-chunk denoising, reducing the pressure to model visual evolution.
We introduce Future Chunk Prediction (FCP), an auxiliary objective jointly optimized with teacher-forced denoising that predicts temporally separated chunks without access to the intervening visual context.
Drawing supervision directly from training videos, FCP requires no additional pretrained visual encoder and leaves autoregressive inference unchanged.
Our method achieves superior physical plausibility over standard teacher-forced autoregressive training with a lower overall training compute budget, highlighting the effectiveness and efficiency of future-chunk supervision for modeling visual evolution.
Intriguingly, our visualizations suggest that FCP fosters a more structured representation space, which may be valuable for building autoregressive world models.
\end{abstract}

\section{Introduction}
\label{sec:introduction}

Autoregressive video generation \citep{LongLive,Self-Forcing,Causal-Forcing,Helios} has emerged as an important direction in video generation, offering a flexible paradigm in which visual sequences can be extended continuously.
It typically operates through chunk-wise temporal rollout, where each new chunk is generated conditioned on preceding visual context.
As an important pathway toward generative world modeling \citep{Genie,GameNGen}, such a history-conditioned generator is expected not merely to produce plausible successive chunks, but also to model how visual scenes evolve over time.

However, the ability to extend a video does not by itself guarantee that the model has learned how the scene should evolve.
We argue that a key reason is standard teacher forcing, where the current chunk is denoised given ground-truth history.
Due to the strong temporal redundancy in videos, neighboring chunks often share highly similar appearance and motion patterns. 
As a result, the model may rely heavily on local visual continuity to recover the current chunk, creating a shortcut that provides limited pressure to capture the underlying dynamics of scene evolution.

This motivates us to seek supervision that goes beyond what can be readily inferred from local visual continuity.
A natural place to look is farther into the future.
As the prediction target moves farther from the observed history, local appearance and motion continuity provide less direct support for recovering the target, thereby placing greater pressure on the model to capture temporal progression.
Importantly, this supervision comes directly from the training video itself, without requiring external pretrained representations or semantic alignment \citep{REPA,VideoREPA}.

Motivated by this intuition, we introduce Future Chunk Prediction (FCP), a simple auxiliary objective for autoregressive video diffusion training.
In addition to denoising the target chunk under standard teacher forcing, FCP uses the learned intermediate representation to predict a target beyond the current chunk, without exposing the model to the intervening ground-truth visual context.
This complementary objective places additional training pressure on the learned representations to capture information relevant to visual evolution, a capability fundamental to effective temporal rollout in autoregressive video generation.
Importantly, FCP is introduced only during training and leaves the original autoregressive inference process unchanged.

To assess whether this additional supervision improves the modeling of visual evolution, we focus on physical plausibility as a concrete and challenging evaluation dimension.
Our method demonstrates improved physical plausibility over standard teacher-forced autoregressive training under matched data exposure and training budget, highlighting the effectiveness and efficiency of future-chunk supervision.
These results highlight the potential of exploiting richer predictive supervision from existing video data, rather than relying solely on increased training resources.

Beyond the improvements in generation quality, we further examine how FCP affects the intermediate video representations.
Interestingly, the introduced supervision induces more pronounced spatiotemporal organization in the representation space, suggesting that FCP also meaningfully reshapes the representations learned during training.
This finding highlights the potential of future prediction for learning representations better suited to autoregressive video generation, which may further benefit generative world modeling.

\section{Related Work}
\label{sec:related-work}

\subsection{Autoregressive Video Diffusion}
\label{sec:related-ar-video-diffusion}

Conventional video diffusion models typically synthesize fixed-duration clips in a single denoising process \citep{VDM,StableVideoDiffusion,HunyuanVideo-1.5,Wan2025}.
Autoregressive video diffusion instead extends video chunk by chunk, conditioning on preceding visual context \citep{MCVD,Diffusion-Forcing,MAGI-1}.
This formulation supports streaming generation and provides a basis for interactive visual simulation and generative world modeling \citep{Genie,Matrix-Game-2.0,LongLive,GameNGen}.

Prior work addresses rollout quality by adapting bidirectional models into causal few-step generators through distribution matching or adversarial training \citep{CausVid,AAPT}, exposing models to self-generated histories or extended student rollouts \citep{Self-Forcing,Self-Forcing++,Rolling-Forcing}, and examining the causal compatibility of teacher supervision with student generation \citep{Causal-Forcing,Causal-rCM}.
Other approaches manage historical context through active windows, cached states, or global anchors \citep{Ca2-VDM,LongLive,Rolling-Forcing}, while selective retrieval and cache compression balance long-range consistency against computational and memory costs \citep{MoC,TempCache,Forcing-KV}.
Our approach complements these directions by adding future-chunk supervision to teacher-forced training without prescribing rollout distillation or inference-time context management.

\subsection{Representation Alignment for Generation}
\label{sec:related-representation-alignment}

Aligning intermediate diffusion features with vision foundation models can accelerate training and improve generation quality \citep{REPA}.
Video extensions use cross-frame and multi-feature alignment \citep{CREPA,Align4Gen}, or relational supervision targeting physical plausibility \citep{VideoREPA}.
These methods rely on targets from external pretrained encoders \citep{DINOv2,VideoMAEv2}, which may complicate scaling to large video corpora.
Our approach instead derives supervision from future chunks in the training videos, without an additional pretrained vision model, offering a data-native path toward scaling.

Our work was developed independently of and concurrently with Next Forcing \citep{Next-Forcing}, the closest related approach. We include relevant comparisons in our experiments and provide a more detailed discussion in the appendix.

\section{Preliminaries}
\label{sec:preliminaries}

\subsection{Flow Matching}
\label{sec:flow-matching}

Flow matching learns a time-dependent vector field that continuously connects the data distribution to a simple noise distribution.
Let $\mathbf{x}$ denote a clean video sample, $\boldsymbol{\epsilon}\sim\mathcal{N}(\mathbf{0},\mathbf{I})$ denote Gaussian noise, and $t\sim\mathcal{U}(0,1)$.
We define the intermediate state along a linear probability path as
\begin{equation}
  \widetilde{\mathbf{x}}_t
  = (1-t)\mathbf{x} + t\boldsymbol{\epsilon},
\end{equation}
whose target velocity is $\boldsymbol{\epsilon}-\mathbf{x}$.
A conditional velocity model $\mathbf{v}_{\theta}$ is trained with
\begin{equation}
  \mathcal{L}_{\mathrm{FM}}(\theta)
  = \mathbb{E}_{\mathbf{x},\boldsymbol{\epsilon},t,\mathbf{c}}
  \left[
    \left\lVert
      \mathbf{v}_{\theta}(\widetilde{\mathbf{x}}_t,t,\mathbf{c})
      - (\boldsymbol{\epsilon}-\mathbf{x})
    \right\rVert_2^2
  \right],
\end{equation}
where $\mathbf{c}$ denotes optional conditioning information.
At inference, a numerical sampler follows the learned velocity field to transform noise at $t=1$ into a video sample at $t=0$.

\subsection{Autoregressive Video Generation}
\label{sec:autoregressive-video-generation}

Autoregressive video generation, also referred to as streaming video generation, generates videos through chunk-wise temporal rollout \citep{CausVid}.
In diffusion-based instantiations, each autoregressive step generates a multi-frame chunk conditioned on the preceding observed or generated context.
Repeatedly applying a fixed-length chunk generator supports long-form video generation over a flexible number of chunks.
This chunk-wise formulation is distinct from discrete-token autoregressive video generation \citep{CogVideo,MAGVIT-v2,VideoPoet,EMU3}, which sequentially predicts tokenized representations, and from single-shot generation, which produces a fixed-length clip in one invocation.

Let $\mathbf{x}_{1:K}=\{\mathbf{x}_1,\ldots,\mathbf{x}_K\}$ denote a video divided into $K$ ordered, non-overlapping chunks of equal length, and let $\mathbf{c}$ denote the text condition.
The conditional distribution of the complete video can be factorized as
\begin{equation}
  p(\mathbf{x}_{1:K}\mid\mathbf{c})
  = \prod_{k=1}^{K} p(\mathbf{x}_k\mid\mathbf{x}_{<k},\mathbf{c}),
  \label{eq:ar-factorization}
\end{equation}
where $\mathbf{x}_{<k}$ denotes the chunks preceding $\mathbf{x}_k$.

\subsection{Teacher-Forced Training}
\label{sec:teacher-forcing}

Different training paradigms for autoregressive video generation construct the historical context using either clean chunks \citep{MCVD,Self-Forcing,GPDiT} or noise-perturbed ones \citep{Diffusion-Forcing,Rolling-Diffusion}.
Here, we adopt the clean-history setting.
A common realization of this setting is teacher-forced training, in which the historical context $\mathbf{x}_{<k}$ is taken directly from the ground-truth video.
Under the flow-matching formulation introduced in Section~\ref{sec:flow-matching}, Gaussian noise $\boldsymbol{\epsilon}_k$ and a flow time $t$ are sampled to perturb the target chunk $\mathbf{x}_k$ along the linear probability path, yielding $\widetilde{\mathbf{x}}_{k,t}$.
The model predicts the corresponding velocity while conditioned on the clean preceding chunks $\mathbf{x}_{<k}$ and text condition $\mathbf{c}$:
\begin{equation}
  \mathcal{L}_{\mathrm{TF}}(\theta)
  = \mathbb{E}_{\mathbf{x}_{1:K},\mathbf{c},k,t,\boldsymbol{\epsilon}_k}
  \left[
    \left\lVert
      \mathbf{v}_{\theta}
      (\widetilde{\mathbf{x}}_{k,t},t,\mathbf{x}_{<k},\mathbf{c})
      - (\boldsymbol{\epsilon}_k-\mathbf{x}_k)
    \right\rVert_2^2
  \right].
\end{equation}
Together, this defines a teacher-forced flow-matching objective for chunk-wise autoregressive video generation.

\section{Method}
\label{sec:method}

\begin{figure}[th]
  \centering
  \includegraphics[width=0.9\linewidth]{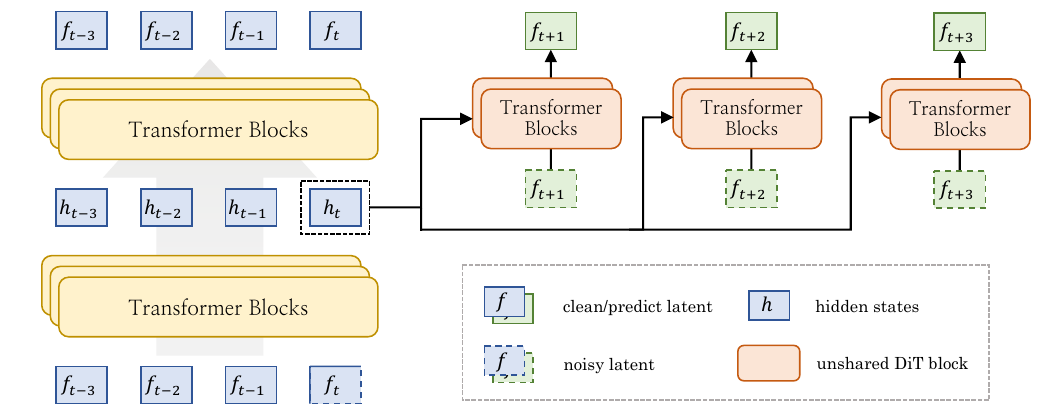}
  \caption{\textbf{Overview of Future Chunk Prediction (FCP).} Intermediate representations from the autoregressive backbone condition auxiliary transformer blocks for future-chunk prediction.}
  \label{fig:fcp-overview}
\end{figure}

In this section, we detail our proposed method.
Section~\ref{sec:fcp} introduces the general formulation of Future Chunk Prediction (FCP) and formalizes its auxiliary future-prediction objective, while Section~\ref{sec:fcp-architecture} presents the FCP architecture adopted in this work.

\subsection{Future Chunk Prediction}
\label{sec:fcp}

Building upon the teacher-forced formulation in Section~\ref{sec:teacher-forcing}, each target chunk is denoised conditioned on its temporally contiguous ground-truth history.
As discussed in Section~\ref{sec:introduction}, the strong temporal redundancy of video allows this immediately preceding correct context to provide substantial support for current-chunk denoising, reducing the pressure to capture temporal progression and potentially hindering the learning of representations that capture visual evolution.
This observation suggests a simple design principle: we seek a learning signal that is less directly supported by such immediately preceding context, yet remains intrinsic to the training video itself, allowing the supervision to scale naturally with video data without relying on external pretrained representations.

To this end, we introduce Future Chunk Prediction (FCP), which supplements next-chunk denoising with an auxiliary task of predicting temporally more distant chunks from the same video.
By excluding the intervening chunks from the visual context, this task discourages reliance on local temporal continuity and encourages stronger predictive representations of visual evolution.

Formally, re-indexing a factor in \eqref{eq:ar-factorization}, standard next-chunk prediction given the observed prefix $\mathbf{x}_{\le k}$ models
\[
  p(\mathbf{x}_{k+1}\mid\mathbf{x}_{\le k},\mathbf{c}).
\]
FCP instead requires the same available context to support prediction of a farther target:
\begin{equation}
  p_{\mathrm{FCP}}(\mathbf{x}_{k+\Delta}\mid\mathbf{x}_{\le k},\mathbf{c}),
  \qquad \Delta>1,\quad k+\Delta\le K,
  \label{eq:fcp-prediction}
\end{equation}
where $\Delta$ is an integer temporal offset measured in chunks.
In this formulation, the intervening chunks $\mathbf{x}_{k+1:k+\Delta-1}$ are excluded from the conditioning context, yielding a supervisory signal that is distinct from and complementary to standard teacher forcing.

A natural implementation of FCP uses an auxiliary predictor with dedicated parameters.
In preliminary experiments, sharing generator parameters between teacher-forced denoising and masked-context future prediction degraded temporal coherence, possibly because the two objectives place competing demands on shared parameters.

However, training this predictor in isolation would leave the main generator unaffected.
We therefore condition the predictor on the generator's intermediate hidden states and backpropagate the auxiliary loss through them \citep{DeepSeek-V3}.
This design regularizes these representations to capture visual evolution and better support autoregressive rollout.

For a practical realization of the above formulation, let $\mathbf{h}_k$ denote the generator's intermediate hidden states corresponding to chunk $k$.
The auxiliary prediction is then expressed as
\begin{equation}
  p_{\mathrm{FCP}}\!\left(\mathbf{x}_{k+\Delta}\mid\mathbf{h}_k\right),
  \qquad \Delta\ge1.
  \label{eq:fcp-hidden-prediction}
\end{equation}

Compared with \eqref{eq:fcp-prediction}, this formulation incorporates two deliberate choices regarding the conditioning information.
First, we omit explicit text conditioning, encouraging the auxiliary predictor to derive predictive information from $\mathbf{h}_k$ rather than a separate textual input.
Second, we condition the auxiliary predictor on $\mathbf{h}_k$ rather than $\mathbf{h}_{\le k}$, encouraging a more compact encoding of predictive information while reducing computational cost.
We examine the effects of these conditioning choices through ablation studies in Section~\ref{sec:experiments}.

\subsection{FCP Architecture}
\label{sec:fcp-architecture}

Building on the formulation in \eqref{eq:fcp-hidden-prediction}, we present one concrete implementation of FCP to evaluate the effectiveness of the proposed approach.
For the autoregressive generator, we use a standard video diffusion model \citep{Wan2025} whose backbone consists of stacked diffusion transformer (DiT) blocks with chunk-causal attention.
The auxiliary predictor adopts a lightweight counterpart of this DiT architecture with a separate set of parameters.
Figure~\ref{fig:fcp-overview} illustrates the overall architecture, with component-specific designs described below.

\paragraph{Source Representations.}
We extract intermediate features associated with chunk $k$ in both clean and noise-perturbed states, denoted by $\mathbf{h}_k^{\mathrm{clean}}$ and $\mathbf{h}_k^{\mathrm{noisy}}$, respectively.
Clean context provides historical information for subsequent chunk generation, while the noisy target is progressively denoised to produce the current chunk.
We consider both roles essential to autoregressive generation and therefore regularize both types of representations through future-prediction supervision.

\begin{figure}[t]
  \centering
  \captionsetup[subfigure]{font=normalsize,justification=justified,singlelinecheck=false,skip=3pt,hypcap=true}
  \begin{subfigure}[t]{0.49\linewidth}
    {\centering
    \includegraphics[height=0.41\linewidth]{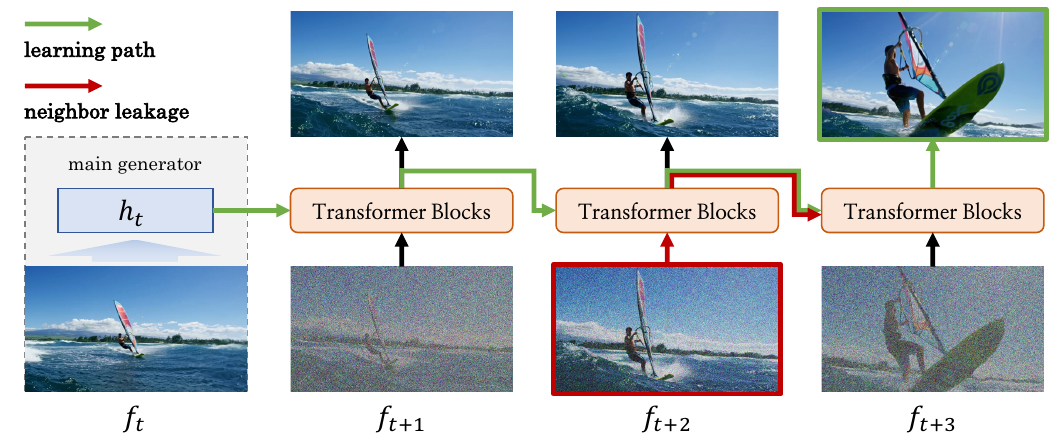}\par}
    \caption{Chained prediction: information from noisy intermediate targets may leak to subsequent predictors along the \textcolor{red!70!black}{red} arrows.}
    \label{fig:fcp-chained}
  \end{subfigure}\hfill
  \begin{subfigure}[t]{0.49\linewidth}
    {\centering
    \includegraphics[height=0.41\linewidth]{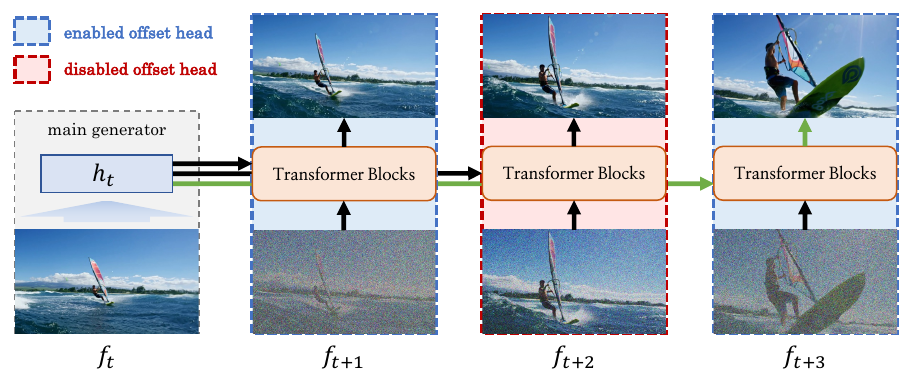}\par}
    \caption{Parallel prediction: offset-specific predictors can run concurrently across GPUs and be flexibly enabled or disabled during training.}
    \label{fig:fcp-parallel}
  \end{subfigure}
  \caption{Two variants of future-prediction supervision, illustrated through the \textcolor{green!60!black}{green} paths from $\mathbf{h}_t$ to $f_{t+3}$. The chained design passes hidden states between predictors, whereas the parallel design conditions each predictor directly on the source representations.}
  \label{fig:parallel-chained-prediction}
\end{figure}

\paragraph{Parallel Future Prediction.}
Each future offset has a separate predictor conditioned directly on $\mathbf{h}_k$, rather than on hidden states from preceding predictors \citep{DeepSeek-V3,Next-Forcing}.
In the chained design (Figure~\ref{fig:fcp-chained}), information from the noisy input for $f_{t+2}$ can reach the predictor for $f_{t+3}$ along the red path.
Given video's temporal redundancy, this may create a shortcut that weakens the pressure on the source representation to encode future evolution.
Our parallel design (Figure~\ref{fig:fcp-parallel}) removes inter-offset information flow, preventing this particular shortcut.
Its independent branches could also run concurrently across GPUs or be flexibly loaded and unloaded during training.

\paragraph{Feature Conditioning.}
To capture information at different semantic levels, we extract features from multiple generator layers, concatenate them along the channel dimension at corresponding token positions, and fuse them with an MLP.
Inspired by prior AdaLN-based conditioning designs \citep{DDT,RepFusion,UniDDT}, we inject the fused features into every auxiliary DiT block through token-wise AdaLN modulation, preserving fine-grained conditioning across spatiotemporal positions.

\paragraph{Asymmetric Text Conditioning.}
Unlike the text-conditioned main generator, the auxiliary predictors omit text cross-attention from every DiT block.
Consistent with \eqref{eq:fcp-hidden-prediction}, this asymmetry blocks a direct pathway through which text prompts could leak cues about future events, encouraging the predictors to rely on the generator's source representations instead.
This design also reduces the predictors' parameter count and computational cost.

\subsection{Training and Inference}
\label{sec:training-inference}

\paragraph{Training.}
We jointly optimize the autoregressive generator and auxiliary predictors using flow matching.
For each offset $\Delta$ in a predefined set $\mathcal{D}$, a separate auxiliary predictor follows the flow-matching objective in Section~\ref{sec:flow-matching}, conditioned on either clean or noisy source representations.
Target noise and flow time are resampled for each auxiliary prediction.
Let $\mathcal{L}_\Delta^{\mathrm{clean}}$ and $\mathcal{L}_\Delta^{\mathrm{noisy}}$ denote the corresponding losses, each averaged over all valid supervision pairs $(k,k+\Delta)$ satisfying $k+\Delta\le K$.
The resulting joint objective takes the form:
\begin{equation}
  \mathcal{L}_{\mathrm{total}}
  = \mathcal{L}_{\mathrm{TF}}
  + \sum_{\Delta\in\mathcal{D}} w_\Delta
  \frac{
    \mathcal{L}_\Delta^{\mathrm{clean}}
    + \mathcal{L}_\Delta^{\mathrm{noisy}}
  }{2}.
  \label{eq:fcp-joint-objective}
\end{equation}
Here, $w_\Delta$ controls the contribution of each offset.
Clean and noisy source representations share the predictor for a given offset, and the auxiliary losses backpropagate through these representations into the generator.

\paragraph{Inference.}
At inference, we discard all FCP-specific modules, retaining the same autoregressive generator and rollout procedure as the teacher-forcing baseline without additional overhead.
Qualitative visualizations of the auxiliary predictors suggest that they capture plausible future evolution despite limited visual fidelity.
We leave their potential use for sampling acceleration through early-stage denoising or generation enhancement through prediction ensembling to future work.

\section{Experiments}
\label{sec:experiments}
\definecolor{physicsiqscore}{RGB}{238, 238, 238}
\definecolor{metricgain}{RGB}{0, 120, 70}
\definecolor{metricdrop}{RGB}{180, 45, 45}

\subsection{Implementation Details}
\label{sec:implementation-details}

We initialize our autoregressive generators from official Wan checkpoints \citep{Wan2025}.
For our implementation, we choose the image-to-video setting, treating the initial frame as fixed clean context and generating subsequent chunks of three latent frames.
Each auxiliary predictor contains three DiT blocks initialized from the corresponding generator; the generators and predictors are then jointly optimized.
Further model and training details are provided in Appendix~\ref{app:experimental-details}.

\subsection{Benchmarks}
\label{sec:benchmarks}

Modeling visual evolution involves diverse aspects of temporal dynamics, making its comprehensive evaluation an open challenge \citep{VSTAR,STEVO-Bench,WorldReasonBench}.
In this work, we focus primarily on physical plausibility \citep{Sora,PhyWorld,Cosmos}, motivated by the availability of dedicated benchmarks and systematic research on its evaluation.
This choice is further motivated by the role of physical plausibility in shaping the perceived realism of generated videos.
We adopt Physics-IQ Verified \citep{Physics-IQ-Verified} as our primary benchmark because its reference-based numerical metrics assess generated continuations against real-world recordings without relying on VLM-based judgments.
We further evaluate on WorldModelBench \citep{WorldModelBench}, PAI-Bench-G \citep{PAI-Bench}, and PhyGround \citep{PhyGround} as complementary physics-related benchmarks.

\begin{table}[!t]
  \caption{
  Physics-IQ Verified results for image-to-video generation \citep{Physics-IQ-Verified}.
  Baseline scores are primarily from the benchmark.
  IQ is the overall score.
  Params counts visual-denoiser parameters.
  \textbf{Bold} and \underline{underlined} mark the best and second-best autoregressive scores.
  }
  \label{tab:physics-iq-sota}
  \begin{center}
  \begin{tabular}{lcccccc}
    \toprule
    \textbf{Model}
      & \textbf{Params}
      & \textbf{IQ $\uparrow$}
      & \textbf{SP $\uparrow$}
      & \textbf{ST $\uparrow$}
      & \textbf{WS $\uparrow$}
      & \textbf{MSE $\uparrow$} \\
    \midrule
    \multicolumn{7}{l}{\footnotesize\textcolor{gray}{Bidirectional}} \\
    Wan2.2-5B \citep{Wan2025}
      & 5B & \cellcolor{physicsiqscore}24.9 & 34.2 & 22.1 & 21.1 & 22.1 \\
    Sora 2 \citep{Sora2}
      & -- & \cellcolor{physicsiqscore}26.5 & 37.3 & 27.0 & 26.9 & 14.8 \\
    Cosmos3-Nano \citep{Cosmos3}
      & A8B & \cellcolor{physicsiqscore}29.1 & 40.4 & 22.0 & 24.6 & 29.5 \\
    Wan2.2-14B \citep{Wan2025}
      & A14B & \cellcolor{physicsiqscore}32.2 & 51.1 & 20.5 & 28.5 & 28.9 \\
    HunyuanVideo-1.5 \citep{HunyuanVideo-1.5}
      & 8.3B & \cellcolor{physicsiqscore}33.4 & 47.1 & 26.9 & 29.7 & 30.0 \\
    Grok Imagine Video \citep{GrokImagineVideo}
      & -- & \cellcolor{physicsiqscore}34.8 & 52.7 & 21.4 & 35.7 & 29.6 \\
    \midrule
    \multicolumn{7}{l}{\footnotesize\textcolor{gray}{Autoregressive}} \\
    Causal Forcing \citep{Causal-Forcing}
      & 1.3B & \cellcolor{physicsiqscore}12.1 & 21.4 & 8.5 & 11.4 & 7.2 \\
    Causal Forcing\ \textsubscript{\scriptsize Stage 1} \citep{Causal-Forcing}
      & 1.3B & \cellcolor{physicsiqscore}19.2 & 26.7 & 15.4 & 16.9 & 17.9 \\
    Causal Forcing++ \citep{Causal-Forcing++}
      & 1.3B & \cellcolor{physicsiqscore}21.8 & 31.9 & 20.8 & 18.9 & 15.6 \\
    Helios-Base \citep{Helios}
      & 14B & \cellcolor{physicsiqscore}26.9 & 32.7 & 31.7 & 21.8 & 21.4 \\
    MAGI-1-4.5B \citep{MAGI-1}
      & 4.5B & \cellcolor{physicsiqscore}\underline{31.1} & 27.6 & \textbf{49.5} & 20.8 & 26.6 \\
    \textbf{Wan2.1-1.3B + FCP (Ours)}
      & 1.3B
      & \cellcolor{physicsiqscore}30.9
      & \underline{36.2}
      & 29.6
      & \underline{23.2}
      & \textbf{34.5} \\
    \textbf{Wan2.2-5B + FCP (Ours)}
      & 5B
      & \cellcolor{physicsiqscore}\textbf{33.8}
      & \textbf{36.8}
      & \underline{41.1}
      & \textbf{25.2}
      & \underline{31.9} \\
    \bottomrule
  \end{tabular}
  \end{center}
\end{table}

\subsection{Datasets}
\label{sec:datasets}

We train our models on a mixture of internally sourced videos and WISA-80K \citep{WISA-80K}, using the original text annotations from each source.
The training data primarily comprises general-purpose real-world videos, without using data from physics simulators.
We select videos longer than five seconds and use approximately the first five seconds, sampled at the default frame rate of the corresponding model.
Incorporating physics-simulator data \citep{Cosmos3} may further improve physical plausibility as training scales, but we leave this possibility to future work.

\subsection{Main Results}
\label{sec:main-results}

\begin{table}[!t]
  \caption{
  Controlled Physics-IQ Verified \citep{Physics-IQ-Verified} comparisons on Wan2.2-TI2V-5B.
  The shaded row and columns mark our default configuration and overall IQ scores, respectively.
  }
  \label{tab:wan22-5b-experiments}
  \begin{center}
  \setlength{\tabcolsep}{0.9\tabcolsep}
  \begin{tabular}{clcccccccc}
    \toprule
    \multirow{2}{*}[-\dimexpr(\aboverulesep+\belowrulesep+\cmidrulewidth)/2\relax]{\textbf{ID}}
      & \multirow{2}{*}[-\dimexpr(\aboverulesep+\belowrulesep+\cmidrulewidth)/2\relax]{\textbf{Method}}
      & \multicolumn{4}{c}{\textbf{I2V}}
      & \multicolumn{4}{c}{\textbf{V2V}} \\
    \cmidrule(lr){3-6}\cmidrule(lr){7-10}
      & & \textbf{IQ $\uparrow$}
      & \textbf{SP $\uparrow$}
      & \textbf{ST $\uparrow$}
      & \textbf{WS $\uparrow$}
      & \textbf{IQ $\uparrow$}
      & \textbf{SP $\uparrow$}
      & \textbf{ST $\uparrow$}
      & \textbf{WS $\uparrow$} \\
    \midrule
    (a) & Teacher Forcing
      & \cellcolor{physicsiqscore}30.4
      & 34.8
      & 32.8
      & 24.0
      & \cellcolor{physicsiqscore}40.6
      & 48.5
      & 40.8
      & 37.8 \\
    (b) & \quad + FCP (3 offsets)
      & \cellcolor{physicsiqscore}32.1
      & 33.6
      & 39.3
      & 23.2
      & \cellcolor{physicsiqscore}43.0
      & 50.3
      & 44.6
      & 39.8 \\
    \textcolor{black!60}{(c)} & \textcolor{black!60}{\quad + FCP (3 offsets, chained)}
      & \cellcolor{physicsiqscore}\textcolor{black!60}{30.7}
      & \textcolor{black!60}{34.6}
      & \textcolor{black!60}{33.0}
      & \textcolor{black!60}{23.3}
      & \cellcolor{physicsiqscore}\textcolor{black!60}{40.5}
      & \textcolor{black!60}{48.5}
      & \textcolor{black!60}{39.3}
      & \textcolor{black!60}{37.9} \\
    \rowcolor{physicsiqscore}
    (d) & \quad + FCP (5 offsets)
      & \cellcolor{physicsiqscore}33.8
      & 36.8
      & 41.1
      & 25.2
      & \cellcolor{physicsiqscore}44.4
      & 51.1
      & 47.2
      & 40.7 \\
    \textcolor{black!60}{(e)} & \textcolor{black!60}{\quad + FCP (5 offsets, chained)}
      & \cellcolor{physicsiqscore}\textcolor{black!60}{31.0}
      & \textcolor{black!60}{35.6}
      & \textcolor{black!60}{33.0}
      & \textcolor{black!60}{23.9}
      & \cellcolor{physicsiqscore}\textcolor{black!60}{39.9}
      & \textcolor{black!60}{47.6}
      & \textcolor{black!60}{38.9}
      & \textcolor{black!60}{36.6} \\
    (f) & \quad + FCP (offsets 1, 3, 5)
      & \cellcolor{physicsiqscore}33.1
      & 36.9
      & 38.5
      & 25.5
      & \cellcolor{physicsiqscore}43.5
      & 51.8
      & 43.8
      & 40.6 \\
    \bottomrule
  \end{tabular}
  \end{center}
\end{table}

\begin{figure}[!t]
  \centering
  \includegraphics[width=0.495\linewidth]{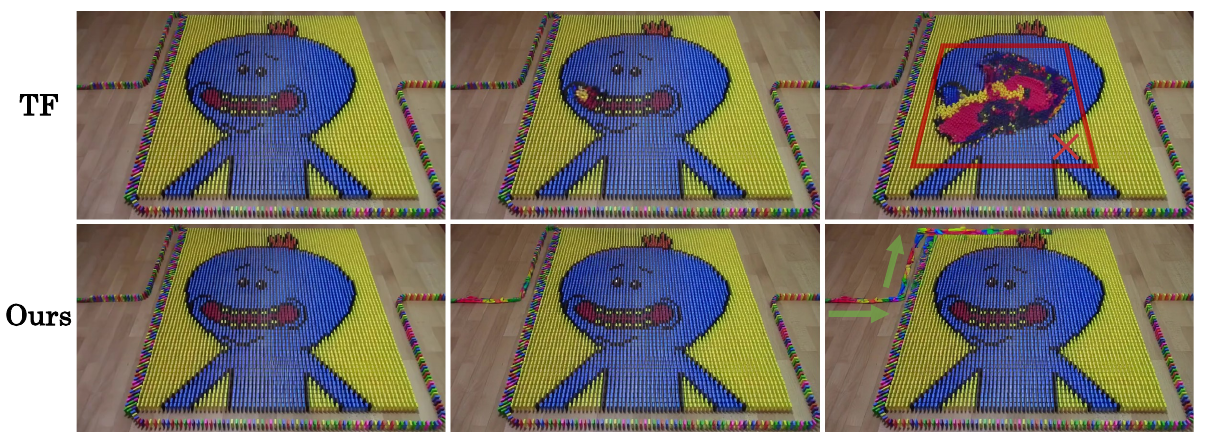}\hfill
  \includegraphics[width=0.495\linewidth]{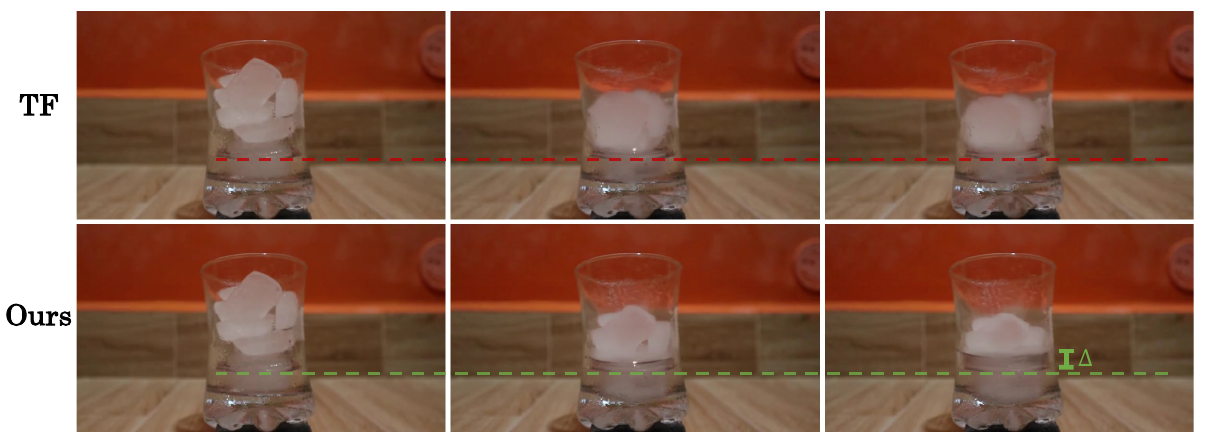}
  \caption{Qualitative comparison of teacher forcing (top) and FCP (bottom).
  Left: FCP predicts domino toppling along the expected direction (green arrows), whereas teacher forcing produces an anomalous collapse at the center (red box).
  Right: FCP captures the rise in water level as the ice melts (green $\Delta$), a detail missed by teacher forcing.}
  \label{fig:physics-qualitative-examples}
\end{figure}

\paragraph{Physics-IQ Verified.}
Differences in training data and settings limit the attribution of cross-model performance gaps to methodological improvements.
We therefore place greater emphasis on the controlled comparisons in Tables~\ref{tab:wan22-5b-experiments} and~\ref{tab:component-ablation}, where matched data and experimental settings enable a fairer assessment.
Compared with teacher forcing, FCP raises the IQ score from 27.6 to 30.9 for the 1.3B model.
For the 5B model, it improves I2V and V2V scores from 30.4 to 33.8 and from 40.6 to 44.4, respectively.
These consistent gains support the effectiveness of future-prediction supervision in improving physical plausibility across scales and conditioning settings.

Table~\ref{tab:physics-iq-sota} provides broader context for these results.
Our 5B model outperforms larger bidirectional models, including Wan2.2-14B \citep{Wan2025}, HunyuanVideo-1.5 \citep{HunyuanVideo-1.5}, and Cosmos3-Nano \citep{Cosmos3}, as well as autoregressive pre-training results from comparably sized MAGI-1-4.5B \citep{MAGI-1} and larger Helios-Base \citep{Helios}.

\paragraph{Qualitative Analysis of Physical Dynamics.}
Figure~\ref{fig:physics-qualitative-examples} complements the quantitative results with two examples of temporal evolution.
FCP preserves the direction of domino toppling and captures the water-level rise visible in the ice-melting sequence, whereas the teacher-forcing baseline exhibits an anomalous central collapse in the former and misses the subtle level change in the latter.
Together, these examples suggest that FCP better captures physically plausible visual evolution, which is central to the perceived realism of generated videos.


\begin{table}[!t]
  \caption{
  Results on WorldModelBench, PAI-Bench-G, and PhyGround.
  All variants are based on Wan2.2-TI2V-5B and use matched training settings.
  }
  \label{tab:worldmodelbench-transposed}
  \begin{center}
  \begin{tabular}{clcc}
    \toprule
    \textbf{Benchmark}
      & \textbf{Metric}
      & \textbf{Teacher Forcing}
      & \textbf{+ FCP (Ours)} \\
    \midrule
    \multirow{4}{*}{\shortstack{\textbf{WorldModelBench}\\\citep{WorldModelBench}}}
      & Instruction $\uparrow$
      & 1.77
      & \textbf{1.81}\,\textcolor{metricgain}{\scriptsize(+0.04)} \\
      & Physical Laws $\uparrow$
      & 4.28
      & \textbf{4.47}\,\textcolor{metricgain}{\scriptsize(+0.19)} \\
      & Common Sense $\uparrow$
      & 1.17
      & \textbf{1.36}\,\textcolor{metricgain}{\scriptsize(+0.19)} \\
    \cmidrule(lr){2-4}
      & \textbf{Total $\uparrow$}
      & 7.22
      & \textbf{7.64}\,\textcolor{metricgain}{\scriptsize(+0.42)} \\
    \midrule
    \multirow{7}{*}{\shortstack{\textbf{PAI-Bench-G}\\\citep{PAI-Bench}}}
      & AV $\uparrow$
      & 63.54
      & \textbf{64.81}\,\textcolor{metricgain}{\scriptsize(+1.27)} \\
      & Common Sense $\uparrow$
      & 90.90
      & \textbf{92.10}\,\textcolor{metricgain}{\scriptsize(+1.20)} \\
      & Human $\uparrow$
      & 81.62
      & \textbf{82.46}\,\textcolor{metricgain}{\scriptsize(+0.84)} \\
      & Industry $\uparrow$
      & 85.65
      & \textbf{87.24}\,\textcolor{metricgain}{\scriptsize(+1.59)} \\
      & Physics $\uparrow$
      & 90.09
      & \textbf{90.29}\,\textcolor{metricgain}{\scriptsize(+0.20)} \\
      & Robot $\uparrow$
      & 70.56
      & \textbf{72.95}\,\textcolor{metricgain}{\scriptsize(+2.39)} \\
    \cmidrule(lr){2-4}
      & \textbf{Domain Score $\uparrow$}
      & 81.13
      & \textbf{82.38}\,\textcolor{metricgain}{\scriptsize(+1.25)} \\
    \midrule
    \multirow{2}{*}{\shortstack{\textbf{PhyGround}\\\citep{PhyGround}}}
      & General $\uparrow$
      & 2.63
      & \textbf{2.66}\,\textcolor{metricgain}{\scriptsize(+0.03)} \\
      & Physics $\uparrow$
      & 2.75
      & \textbf{2.80}\,\textcolor{metricgain}{\scriptsize(+0.05)} \\
    \bottomrule
  \end{tabular}
  \end{center}
\end{table}

\paragraph{Other Physics-Related Benchmarks.}
We further evaluate on physics-related benchmarks, including WorldModelBench \citep{WorldModelBench}, PAI-Bench-G \citep{PAI-Bench}, and PhyGround \citep{PhyGround}.
As shown in Table~\ref{tab:worldmodelbench-transposed}, FCP generally outperforms teacher forcing under matched settings, supporting the generalizability of our approach.
These evaluations rely on VLM-based judgments of specified criteria, which may be less sensitive to subtle differences in the timing and magnitude of physical changes than direct comparison with real-world reference videos.
This limited sensitivity may partly explain the modest gains on some metrics.

\subsection{Ablation Studies}
\label{sec:ablation}

\begin{figure}[!t]
  \begin{minipage}[t]{0.49\linewidth}
  \vspace{0pt}
  \captionsetup{justification=justified,singlelinecheck=false}
  \captionof{table}{Ablation of the main FCP design components. The shaded row and column mark the default setting and overall IQ score, respectively.}
  \label{tab:component-ablation}
  \begin{center}
  \setlength{\tabcolsep}{2.0pt}
  \begin{tabular}{ccccccc}
    \toprule
    \textbf{ID}
      & \textbf{Clean}
      & \textbf{Noisy}
      & \textbf{Text}
      & \textbf{Chain}
      & \textbf{IQ $\uparrow$}
      & \textbf{SP $\uparrow$} \\
    \midrule
    (a) & & & &
      & \cellcolor{physicsiqscore}27.6 & 32.8 \\
    \midrule
    (b) & $\checkmark$ & & &
      & \cellcolor{physicsiqscore}28.8 & 33.0 \\
    (c) & & $\checkmark$ & &
      & \cellcolor{physicsiqscore}29.9 & 34.7 \\
    \rowcolor{physicsiqscore}
    (d) & $\checkmark$ & $\checkmark$ & &
      & \cellcolor{physicsiqscore}30.9 & 36.2 \\
    \midrule
    (e) & $\checkmark$ & $\checkmark$ & $\checkmark$ &
      & \cellcolor{physicsiqscore}29.8 & 34.8 \\
    (f) & $\checkmark$ & $\checkmark$ & & $\checkmark$
      & \cellcolor{physicsiqscore}30.1 & 34.3 \\
    \bottomrule
  \end{tabular}
  \end{center}
  \end{minipage}\hfill
  \begin{minipage}[t]{0.48\linewidth}
  \vspace{0.55\baselineskip}
  \begin{center}
    \includegraphics[width=\linewidth]{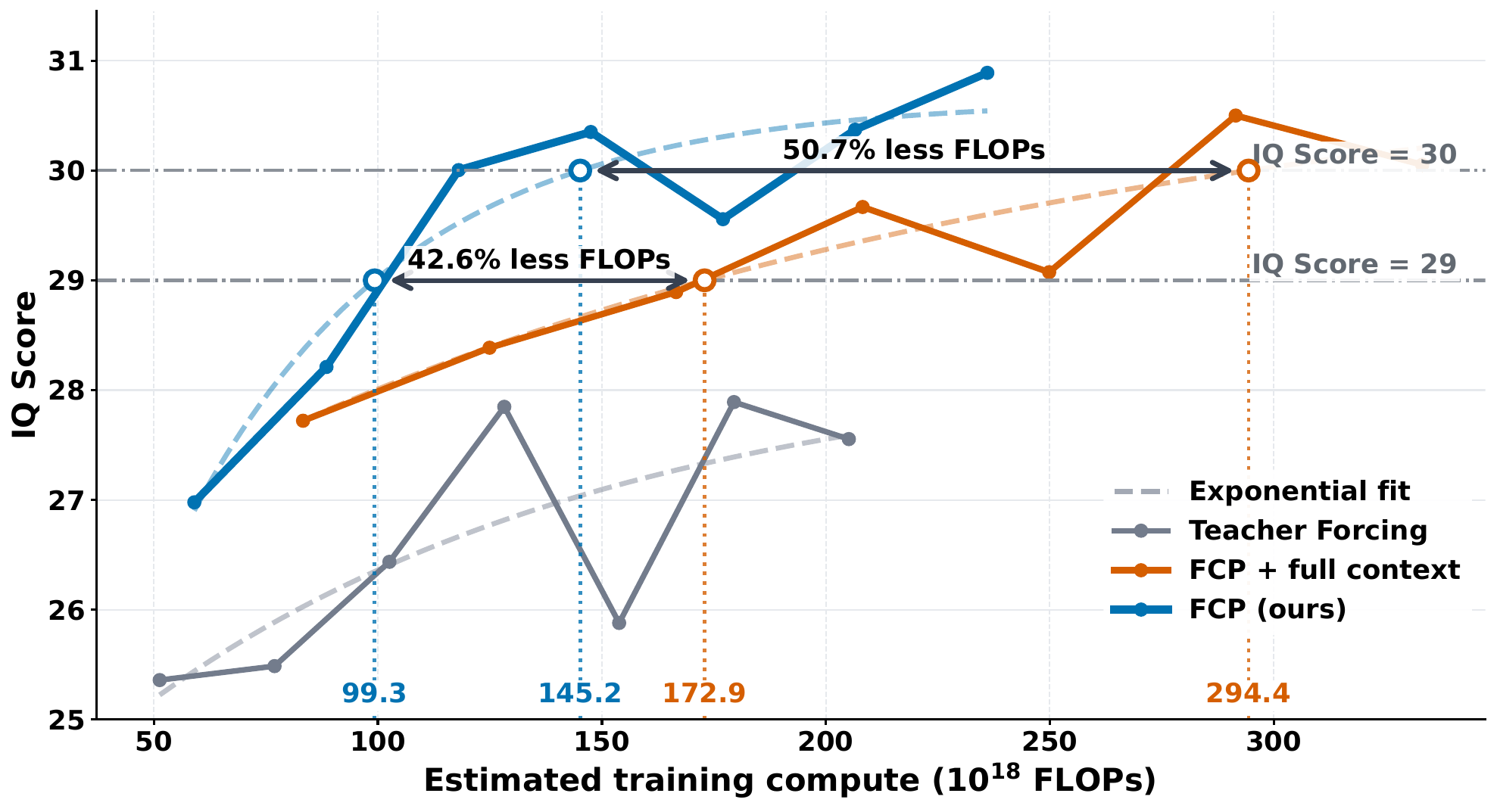}
  \end{center}
  \captionsetup{justification=raggedright,singlelinecheck=false}
  \captionof{figure}{Training convergence on Physics-IQ Verified versus estimated training compute.}
  \label{fig:training-convergence}
  \end{minipage}
\end{figure}

We primarily conduct ablation studies on Wan2.1-1.3B under identical settings across variants; detailed settings and results are provided in Appendices~\ref{app:training-configurations} and~\ref{app:ablation-metrics}, respectively.

\vspace{-8pt}
\paragraph{Source Representations.}
While prior representation-alignment methods \citep{REPA} primarily regularize intermediate features along the denoising process, clean historical context in autoregressive video generation also serves a predictive role, carrying information from the immediate past to support anticipation of future visual evolution.
Results in Table~\ref{tab:component-ablation} support this view: regularizing clean-context or noisy-target representations improves IQ over teacher forcing by 1.2 and 2.3 points, respectively, while their combination yields a 3.3-point gain.

\vspace{-8pt}
\paragraph{Parallel Future Prediction.}
Chained prediction provides a natural alternative \citep{DeepSeek-V3,Next-Forcing} to our parallel design; however, as illustrated in Figure~\ref{fig:fcp-chained}, it may exploit intermediate future information as a shortcut.
Consistent with this concern, Table~\ref{tab:component-ablation} shows that parallel prediction improves the IQ score by 0.8, from 30.1 to 30.9.
The limitation of chained prediction is more pronounced in the controlled 5B comparison in Table~\ref{tab:wan22-5b-experiments}: the two chained variants (3 and 5 offsets) yield only marginal IQ-score gains over teacher forcing on I2V and fail to improve V2V performance.
This larger degradation may arise because the higher-capacity model can more readily exploit the potential leakage of intermediate future information illustrated in Figure~\ref{fig:fcp-chained}.
Together, these results show that our parallel design consistently outperforms the chained alternative across model scales and conditioning settings.
It also supports nonconsecutive future offsets, as evaluated in row~(f) of Table~\ref{tab:wan22-5b-experiments}, through the independent branches shown in Figure~\ref{fig:fcp-parallel}.

\paragraph{Training Efficiency.}
The closely related approach Next Forcing \citep{Next-Forcing} conditions future prediction on the full historical representations $\mathbf{h}_{\le k}$, motivating a full-context FCP variant for comparison.
Figure~\ref{fig:training-convergence} compares teacher forcing, this full-context variant, and our default configuration.
Compared with standard teacher forcing, our configuration achieves higher IQ scores at comparable estimated training compute, indicating that its gains do not arise simply from additional computation.
Furthermore, our configuration is estimated to require 42.6\% and 50.7\% less training compute than the full-context variant to reach the same IQ scores of 29 and 30, respectively.
These results indicate that extending the conditioning context from $\mathbf{h}_k$ to $\mathbf{h}_{\le k}$ provides no additional benefit for modeling video evolution, while the lower computational cost of $\mathbf{h}_k$ makes our formulation better suited to scaling autoregressive video models.

\paragraph{Text-Free Future Prediction.}
Comparing rows (d) and (e) in Table~\ref{tab:component-ablation}, adding explicit text conditioning to the auxiliary predictors reduces the IQ score by 1.1.
This supports withholding text from future predictors, as direct textual cues may reveal future evolution and weaken their reliance on visual representations.

\begin{figure}[!t]
  \begin{minipage}[t]{0.55\linewidth}
    \vspace{0pt}
    \centering
    \includegraphics[height=0.78in]{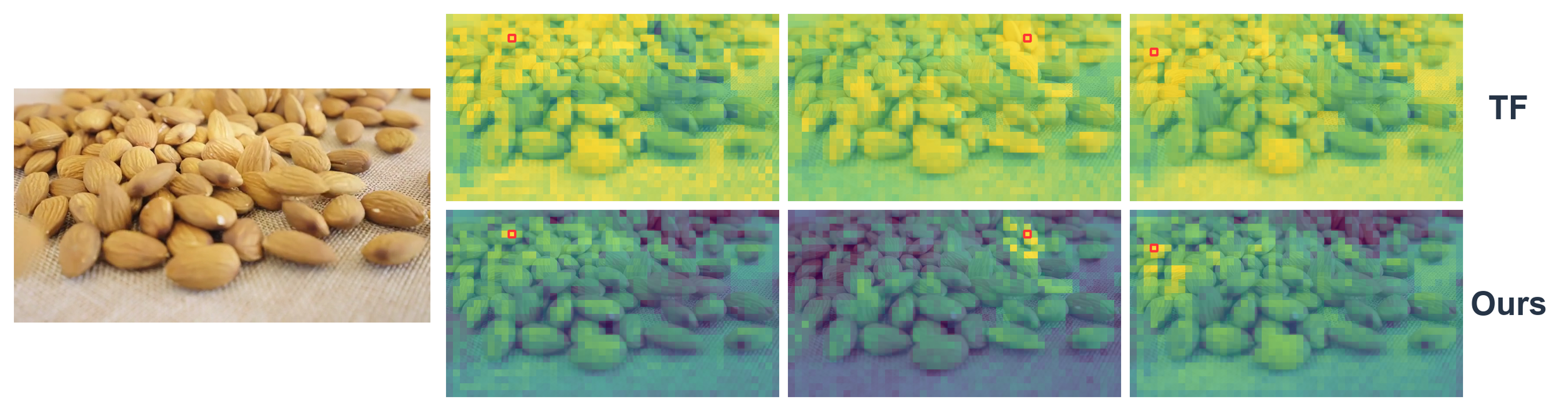}
  \end{minipage}\hfill
  \begin{minipage}[t]{0.43\linewidth}
    \vspace{0pt}
    \centering
    \includegraphics[height=0.78in]{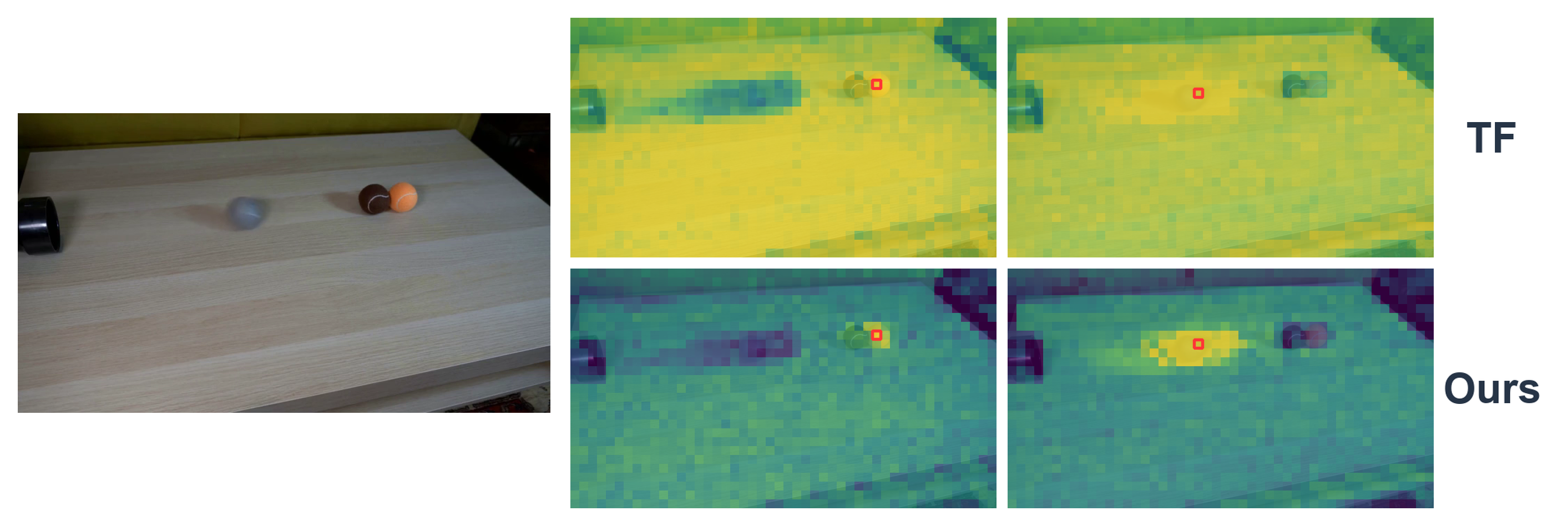}
  \end{minipage}
  \caption{\textbf{Spatial Feature Similarity Maps.} Each map shows within-frame similarity to a selected query token. The top and bottom rows correspond to teacher forcing and FCP, respectively.}
  \label{fig:qualitative-comparison}
\end{figure}

\begin{figure}[!t]
  \centering
  \includegraphics[width=0.49\linewidth]{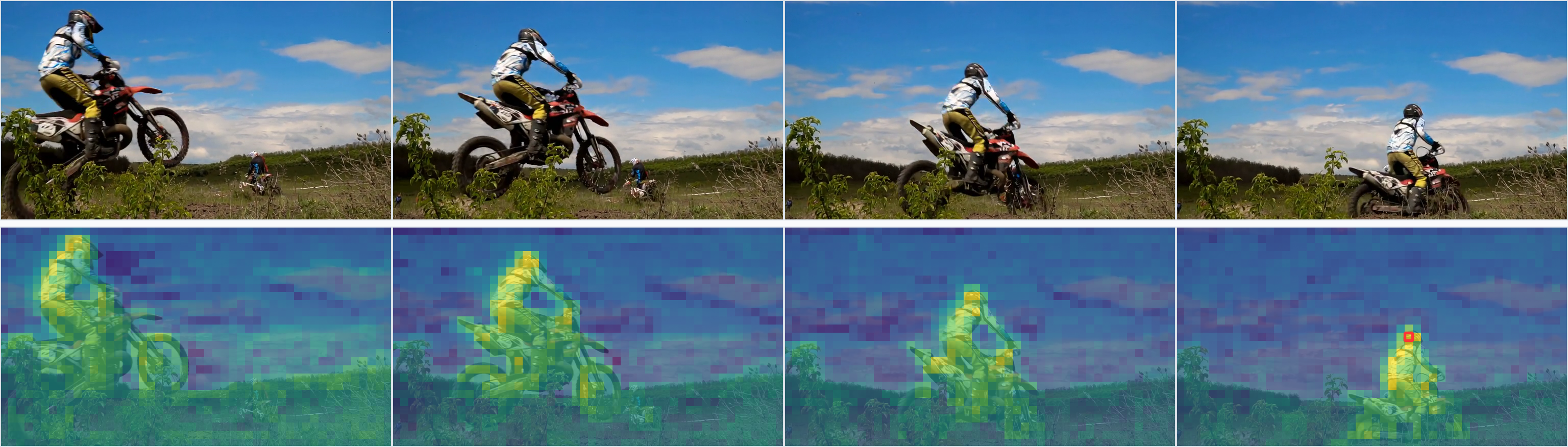}\hfill
  \includegraphics[width=0.49\linewidth]{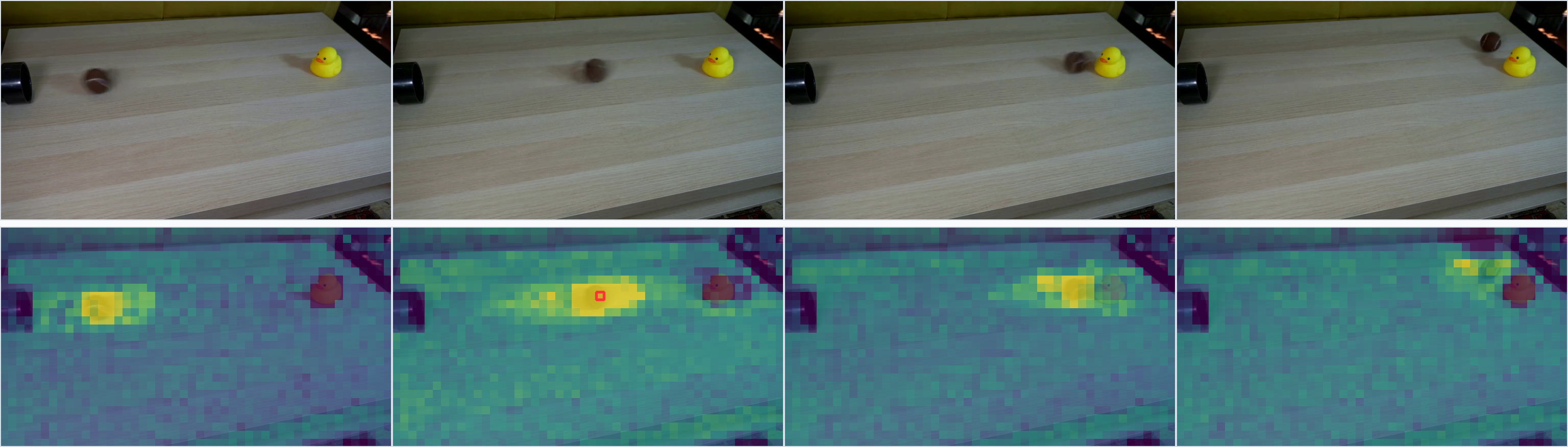}
  \caption{\textbf{Spatiotemporal Feature Similarity Maps.} The top row shows video frames, while the bottom row visualizes similarity to a selected query token across space and time.}
  \label{fig:future-chunk-visualizations}
\end{figure}

\subsection{Representation Space Analysis}
\label{sec:representation-space-analysis}
Beyond physical-plausibility evaluation, we visualize the generator's intermediate representations using query-to-token similarity maps.
As shown in Figure~\ref{fig:qualitative-comparison}, teacher forcing produces diffuse responses extending into unrelated regions, whereas FCP yields more selective responses aligned with the queried object and visually related instances.
Figure~\ref{fig:future-chunk-visualizations} further shows that high-similarity regions remain aligned with corresponding content as it moves across frames.
Together, these visualizations suggest that FCP-trained representations capture clearer spatial structure and meaningful temporal correspondence.
We believe that such semantically structured representations may provide a useful foundation for future world models.

\section{Conclusion}
\label{sec:conclusion}

We introduced Future Chunk Prediction (FCP), an auxiliary objective that predicts temporally separated chunks from the same training video in autoregressive video diffusion.
Optimized jointly with the standard teacher-forced denoising objective, FCP provides complementary supervision to the generator's intermediate representations without requiring an additional pretrained encoder.
The auxiliary predictors are removed at inference, leaving the autoregressive generation process unchanged.
Controlled comparisons show improved physical plausibility across model scales and conditioning settings.
More surprisingly, our visualizations suggest that FCP also induces a more structured representation space, which we believe may be important for autoregressive world models.
Together, these findings support future prediction as a promising training principle for autoregressive video generation.

\section*{Reproducibility Statement}

Sections~\ref{sec:method} and~\ref{sec:experiments} describe our method, training setup, datasets, and evaluation protocols, with further implementation details in Appendix~\ref{app:experimental-details}.
Our code is available at \url{https://github.com/TencentARC/Looking-Ahead}; model checkpoints are not included in this source package.
For the public portion of our training data, we will provide source links and selection criteria; we will also provide representative examples of the internally sourced data.
The complete internal corpus cannot be released, so exact reproduction of training with identical data may not be possible; nevertheless, we will make every effort to support reproducibility.

\section*{AI Use Statement}

Generative AI tools were used to assist with implementing FCP and the associated training, inference, and evaluation code, as well as with translation.
We did not use these tools to generate datasets, develop the conceptual framework, formulate mathematical claims, propose or refine research hypotheses, design or provide feedback on methods or experiments, clean or reformat datasets, conduct qualitative data analysis, or interpret results; this work contains no mathematical proofs.
We also used generative AI for drafting and editing manuscript text, searching for and summarizing relevant literature, writing-related brainstorming, formatting in-text citations, and suggesting paper structure, titles, section headings, and keywords.
The authors reviewed and tested AI-assisted code, checked AI-assisted text and literature recommendations, and take full responsibility for the final content of this work.

\bibliography{references}
\bibliographystyle{arxiv}

\clearpage
\appendix
\makeatletter
\def\fps@table{!htbp}
\makeatother
\section{Additional Experimental Details}
\label{app:experimental-details}

\subsection{Model Configurations}
\label{app:model-configurations}

We initialize the generators from the official Wan \citep{Wan2025} checkpoints listed in Table~\ref{tab:model-configurations}.
For all checkpoints, we implement image-to-video generation by keeping the initial image's latent representation as fixed clean context and generating only subsequent chunks.
Source features are extracted from the outputs of four selected DiT blocks and concatenated at corresponding token positions.
A two-layer MLP with dimensions $4d \rightarrow 2d \rightarrow d$ and a GELU activation fuses these features, where $d$ is the backbone hidden dimension.
Each future offset has an independently parameterized auxiliary branch, including its fusion MLP, patch embedding, DiT blocks, and output head.
Within each offset, all these parameters are shared between the clean and noisy source pathways.
Auxiliary DiT blocks inherit the hidden and FFN dimensions of the corresponding backbone and are initialized from the layers specified in Table~\ref{tab:model-configurations}.
All layer indices are one-based.

\begin{table}
  \caption{Model configurations. All layer indices are one-based. Hidden and FFN dimensions apply to both the backbone and its auxiliary DiT blocks.}
  \label{tab:model-configurations}
  \begin{center}
  \setlength{\tabcolsep}{8pt}
  \begin{tabular}{lcc}
    \toprule
    \textbf{Configuration}
      & \shortstack{\textbf{Wan2.1}\\\textbf{T2V-1.3B}}
      & \shortstack{\textbf{Wan2.2}\\\textbf{TI2V-5B}} \\
    \midrule
    Backbone depth & 30 & 30 \\
    Hidden dimension $d$ & 1536 & 3072 \\
    FFN intermediate dimension & 8960 & 14336 \\
    Source feature layers & 9, 15, 21, 27 & 3, 12, 21, 30 \\
    Auxiliary DiT depth & 3 & 3 \\
    Auxiliary initialization layers & 17, 18, 19 & 28, 29, 30 \\
    Future offsets $\mathcal{D}$ & $\{1,2,3\}$ & $\{1,2,3,4,5\}$ \\
    \bottomrule
  \end{tabular}
  \end{center}
\end{table}

The 1.3B and 5B configurations may appear to follow different design principles.
These differences arise from choices made at different stages of development.

For source-layer selection, the 1.3B configuration evolved from an early single-layer variant using layer 15, subsequently expanded to include both shallower and deeper features.
For the 5B model, we select approximately evenly spaced layers backward from the final block to provide direct auxiliary supervision across the generator's depth, while excluding the earliest blocks whose features may not yet be sufficiently semantically informative.
Computational constraints have prevented a systematic comparison of these source-layer selection strategies, which we may revisit in future scaling efforts.

For auxiliary initialization, the 1.3B configuration follows the early variant using source features from layer 15, with blocks 17--19 used to initialize the auxiliary predictor.
For the 5B model, we instead initialize from the final backbone blocks, decoupling this choice from the source-layer configuration and simplifying subsequent scale-up.
These blocks are also closer to the output and are chosen to better match the output head, which is likewise initialized from the original generator.
Preliminary experiments suggest that this initialization strategy may offer a modest improvement in early-stage convergence.

\subsection{Training Configurations}
\label{app:training-configurations}

\paragraph{Wan2.1-1.3B.}
We use Wan2.1-1.3B for early-stage exploration and controlled ablation studies.
Training runs for 20,000 steps on eight GPUs using 85-frame clips at a resolution of $768 \times 432$ and 16 FPS, with a per-GPU batch size of one and a global batch size of eight.
We use AdamW with $(\beta_1,\beta_2)=(0,0.999)$, a weight decay of 0.01, and a gradient clipping threshold of 0.1.
The generator and auxiliary branches share the same learning rate, which is linearly warmed up to $1 \times 10^{-5}$ over the first 1,000 steps and held constant thereafter.
Both use uniformly sampled flow times followed by a timestep shift of 5.0.
We assign a weight of 1 to the main teacher-forcing loss and set $w_\Delta=0.5$ for every future offset in \eqref{eq:fcp-joint-objective}.
During training, we replace the text prompt with an empty string with a probability of 0.1.
We maintain an exponential moving average of the generator parameters with a decay rate of 0.9995, starting at step 2,000, and use the EMA weights corresponding to each evaluated training step.
All 1.3B ablation experiments follow these training settings.
Unless otherwise specified, the 5B experiments follow the same training settings as the 1.3B model.

\paragraph{Wan2.2-5B.}
The first stage runs for 25,000 steps on 32 GPUs using 121-frame clips at a resolution of $1024 \times 576$ and 24 FPS, with a per-GPU batch size of one and a global batch size of 32.
All other training settings follow the 1.3B configuration.

\subsection{Evaluation Settings}
\label{app:evaluation-settings}

\paragraph{Sampling Settings.}
We use UniPC with 50 denoising steps per chunk, a classifier-free guidance scale of 3.0, and a timestep shift of 5.0 for all models.
Sampling uses the corresponding EMA weights, with all FCP-specific modules removed.
When the inference sequence exceeds the training sequence length, we retain the first three latent frames as attention sinks during KV-cache eviction.

\paragraph{Physics-IQ Verified \citep{Physics-IQ-Verified}.}
We evaluate both image-to-video and video-to-video generation.
For I2V, we condition on the benchmark-provided initial image.
For V2V, we use the longest suffix of the three-second conditioning video compatible with the model's latent and chunk-length constraints.
Both settings generate five-second continuations, and evaluation excludes the conditioning video.
Our models use 16:9 resolutions and require no spatial cropping.

Unless otherwise specified, we use best-practice prompts (BPP) and evaluate all 198 benchmark samples using the official Physics-IQ Verified implementation without modifying its metrics.
Evaluation uses each model's output frame rate specified in Section~\ref{app:training-configurations}, with reference videos at the corresponding frame rate.
We perform four independent runs with seeds 42--45, compute benchmark scores separately for each run, and report their mean.

\paragraph{Other Physics-Related Benchmarks \citep{WorldModelBench,PAI-Bench,PhyGround}.}
We evaluate the teacher-forced and FCP-trained Wan2.2 5B models on all samples of WorldModelBench, PAI-Bench-G, and PhyGround in the I2V setting, using otherwise identical training settings.
We use the benchmark-provided initial images and text prompts, resize images while preserving their aspect ratio, and crop them to the $1024 \times 576$ target resolution when needed.
All three evaluations generate five-second videos at 24 FPS.
For each benchmark, we use its official evaluation code and unmodified metrics, compute scores separately for seeds 42--45, and report their mean in Table~\ref{tab:worldmodelbench-transposed}.

\subsection{Evaluation of Public Models}
\label{app:public-model-evaluation}

Since Physics-IQ Verified serves as our primary benchmark, we supplement its reported comparisons with additional evaluations of relevant public models and settings that are not fully covered in the benchmark paper.
Unless otherwise specified, all evaluations in this subsection follow the common protocol described in Section~\ref{app:evaluation-settings}.

\paragraph{Wan2.2-TI2V-5B \citep{Wan2025}.}
For the official Wan2.2-TI2V-5B checkpoint, we use a resolution of $1280 \times 704$ and consistently crop and align the conditioning images, reference videos, and evaluation masks.
We also tested this checkpoint at the $1024 \times 576$ resolution used by our 5B models, but observed lower scores, potentially reflecting a mismatch with its intended resolution settings.

\paragraph{Causal Forcing and Causal Forcing++ \citep{Causal-Forcing,Causal-Forcing++}.}
We evaluate the official final and Stage 1 checkpoints of Causal Forcing, as well as the official frame-wise 2-step checkpoint of Causal Forcing++, using the repository's inference code and each checkpoint's default sampling settings.
We additionally evaluate Stage 1 because it uses standard teacher-forced training, providing a comparison more closely aligned with our training setting before rollout distillation.
All three checkpoints follow the same preprocessing and evaluation protocol.
For Physics-IQ Verified, we condition on the benchmark-provided initial image and BPP prompt, generating 81 frames at 16 FPS.
We resize while preserving aspect ratio and consistently crop the conditioning images, reference videos, and evaluation masks to $832 \times 480$.
We evaluate all 198 samples with the official metrics, compute scores separately for seeds 42--45, and report their mean.

\paragraph{Helios Base \citep{Helios}.}
We evaluate the official Helios Base checkpoint using the unmodified inference code and repository-default sampling configuration.
We use the officially recommended resolution of $640 \times 384$, applying aspect-ratio-preserving resizing followed by center cropping to the conditioning inputs, reference videos, and evaluation masks with identical transformation parameters.
Helios Base generates 120-frame, five-second continuations at 24 FPS.

\paragraph{MAGI-1-4.5B \citep{MAGI-1}.}
We evaluate the official MAGI-1-4.5B checkpoint using its unmodified inference code and default sampling parameters.
For Physics-IQ Verified, we directly use the benchmark-provided initial images and BPP prompts.
We resize the conditioning images, reference videos, and evaluation masks consistently to $768 \times 432$ without cropping, and generate five-second videos at 24 FPS.
Following the common protocol above, we evaluate all 198 samples with the official Physics-IQ Verified metrics, compute scores separately for seeds 42--45, and report their mean.

\subsection{Representation Visualization Protocol}
\label{app:representation-visualization-protocol}

We use the final 20,000-step EMA checkpoints of the teacher-forced and FCP-trained Wan2.1-1.3B models to visualize their representations on real videos.
Both models receive the same clean video latents, and we extract intermediate features from a single selected source layer of the main generator for each comparison.
The layer and manually selected query-token position are matched between the two models, with the query vector extracted separately from each model's features.
We compute cosine similarity between the query and other tokens in the same frame for spatial maps; for temporal maps, a query from one frame is compared with tokens across the remaining frames.
Similarity values are displayed directly on the token grid without spatial interpolation.
Within each comparison, the teacher-forced and FCP maps use the same color map and a shared value range determined from that comparison's similarity values.

\subsection{Auxiliary Prediction Visualization Protocol}
\label{app:auxiliary-prediction-visualization-protocol}

For Figure~\ref{fig:auxiliary-future-prediction-visualizations}, we use the final 25,000-step EMA checkpoint of our Wan2.2-5B FCP model and visualize predictions on real videos.
The main generator processes clean ground-truth history with its corresponding text prompt, and we extract the hidden states $\mathbf{h}_k$ of the final source chunk.
Conditioned on $\mathbf{h}_k$, each offset-specific auxiliary branch independently samples a future chunk from random noise using the same UniPC settings as standard inference: 50 denoising steps, a classifier-free guidance scale of 3.0, and a timestep shift of 5.0.
The branches receive neither intervening ground-truth or generated chunks nor a separate text input.
For display, we sample decoded frames uniformly and align them temporally with the ground-truth frames.
The auxiliary branches are retained only for this analysis and are otherwise removed during standard inference.

\section{More Results}
\label{app:more-results}

\subsection{Detailed Component Ablations}
\label{app:ablation-metrics}

Table~\ref{tab:component-ablation-full} extends the condensed component ablations in Table~\ref{tab:component-ablation} by reporting the complete breakdown of Physics-IQ Verified metrics, providing a more comprehensive comparison of different component choices.

\begin{table}
  \caption{
  Ablation study of the main components on Physics-IQ Verified \citep{Physics-IQ-Verified}.
  All variants are implemented based on Wan2.1 1.3B.
  The shaded row denotes our default configuration.
  }
  \label{tab:component-ablation-full}
  \begin{center}
  \begin{tabular}{ccccccccc}
    \toprule
    \textbf{clean}
      & \textbf{noisy}
      & \textbf{text}
      & \textbf{chained}
      & \multirow{2}{*}{\textbf{IQ $\uparrow$}}
      & \multirow{2}{*}{\textbf{SP $\uparrow$}}
      & \multirow{2}{*}{\textbf{ST $\uparrow$}}
      & \multirow{2}{*}{\textbf{WS $\uparrow$}}
      & \multirow{2}{*}{\textbf{MSE $\uparrow$}} \\
    \textbf{source}
      & \textbf{source}
      & \textbf{condition}
      & \textbf{connection}
      & & & & & \\
    \midrule
      & & &
      & \cellcolor{physicsiqscore}27.6 & 32.8 & 25.5 & 20.4 & 31.6 \\
    \midrule
    $\checkmark$ & & &
      & \cellcolor{physicsiqscore}28.8 & 33.0 & 29.5 & 21.0 & 31.9 \\
    & $\checkmark$ & &
      & \cellcolor{physicsiqscore}29.9 & 34.7 & 28.2 & 21.6 & 35.0 \\
    \rowcolor{physicsiqscore}
    $\checkmark$ & $\checkmark$ & &
      & \cellcolor{physicsiqscore}30.9 & 36.2 & 29.6 & 23.2 & 34.5 \\
    \midrule
    $\checkmark$ & $\checkmark$ & $\checkmark$ &
      & \cellcolor{physicsiqscore}29.8 & 34.8 & 28.4 & 22.0 & 34.2 \\
    $\checkmark$ & $\checkmark$ & & $\checkmark$
      & \cellcolor{physicsiqscore}30.1 & 34.3 & 29.8 & 21.7 & 34.4 \\
    \bottomrule
  \end{tabular}
  \end{center}
\end{table}

\subsection{Feature Conditioning}
\label{app:feature-conditioning}

Table~\ref{tab:feature-fusion-ablation-full} compares different strategies for injecting source features into the auxiliary predictors.
Concatenating source features along the channel dimension followed by linear fusion is a common and straightforward implementation.
Even with this simple design, FCP improves the IQ score from 27.6 to 29.5 over teacher forcing.
Motivated by prior work demonstrating the effectiveness of AdaLN for feature injection \citep{DDT,RepFusion}, we apply token-wise AdaLN throughout the auxiliary DiT blocks, further increasing the IQ score to 30.9.
We hypothesize that this improvement arises from its layer-wise modulation and learned scaling factors, which provide fine-grained conditioning pathways and strengthen gradient propagation through the source representations into the generator.
Notably, directly concatenating source and future tokens within self-attention is less effective, reaching an IQ score of 28.8.
We hypothesize that joint self-attention weakens the fine-grained correspondence between source and future representations.

\begin{table}
  \caption{
  Ablation study of feature-fusion strategies for injecting features into the future-prediction branch on Physics-IQ Verified \citep{Physics-IQ-Verified}.
  All variants are implemented based on Wan2.1 1.3B.
  The shaded row denotes our default configuration.
  }
  \label{tab:feature-fusion-ablation-full}
  \begin{center}
  \begin{tabular}{lccccc}
    \toprule
    \textbf{Fusion strategy}
      & \textbf{IQ $\uparrow$}
      & \textbf{SP $\uparrow$}
      & \textbf{ST $\uparrow$}
      & \textbf{WS $\uparrow$}
      & \textbf{MSE $\uparrow$} \\
    \midrule
    Token Concat
      & \cellcolor{physicsiqscore}28.8 & 34.1 & 28.1 & 21.0 & 32.0 \\
    Dim Concat
      & \cellcolor{physicsiqscore}29.5 & 32.2 & 31.2 & 20.6 & 34.1 \\
    \rowcolor{physicsiqscore}
    AdaLN
      & \cellcolor{physicsiqscore}30.9 & 36.2 & 29.6 & 23.2 & 34.5 \\
    \bottomrule
  \end{tabular}
  \end{center}
\end{table}

\FloatBarrier
\subsection{Representation Space Visualization}
\label{app:representation-space-visualization}

\begin{figure}[!ht]
  \centering
  \includegraphics[width=\textwidth]{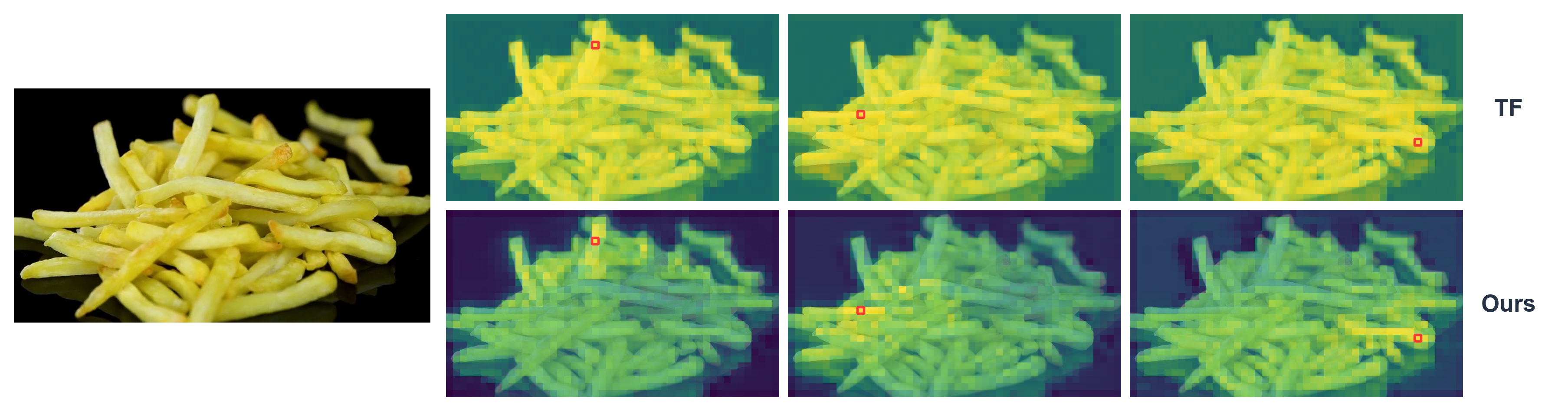}
  \par\smallskip
  \includegraphics[width=\textwidth]{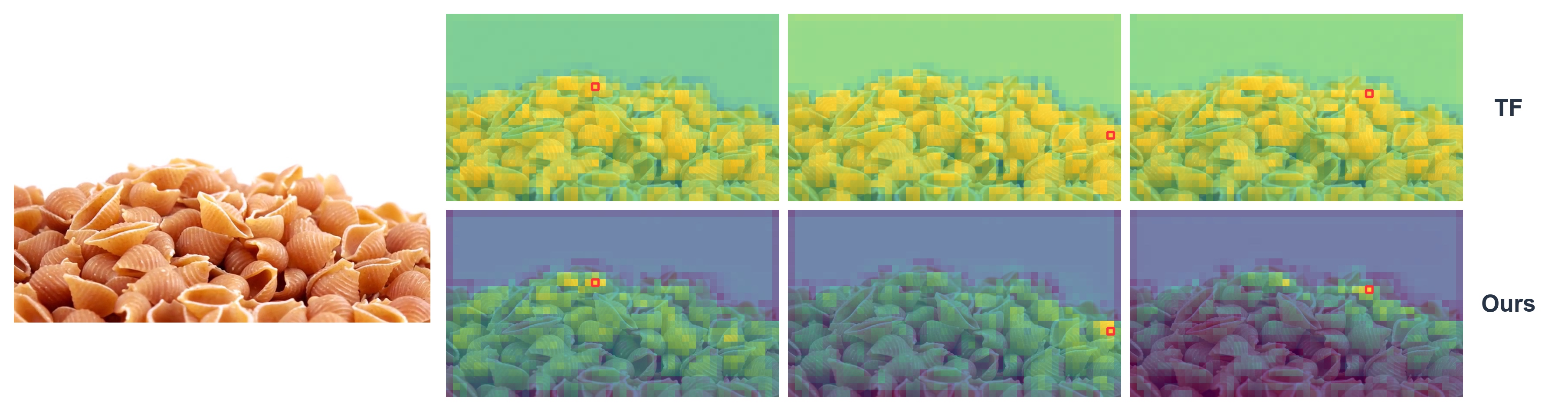}
  \caption{\textbf{Additional Spatial Feature Similarity Maps.}
  Each map shows within-frame similarity to a selected query token; two further examples continue in the next panel.
  Within each example, the top and bottom rows correspond to teacher forcing and FCP, respectively.}
  \label{fig:representation-space-visualizations}
\end{figure}

\begin{figure}[!ht]
  \ContinuedFloat
  \centering
  \includegraphics[width=\textwidth]{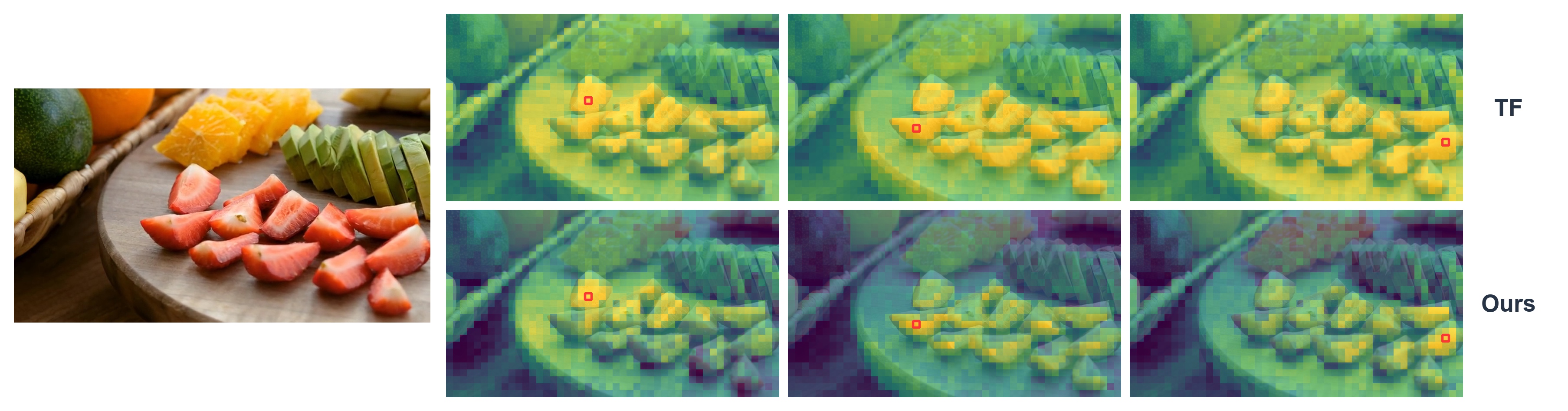}
  \par\smallskip
  \includegraphics[width=\textwidth]{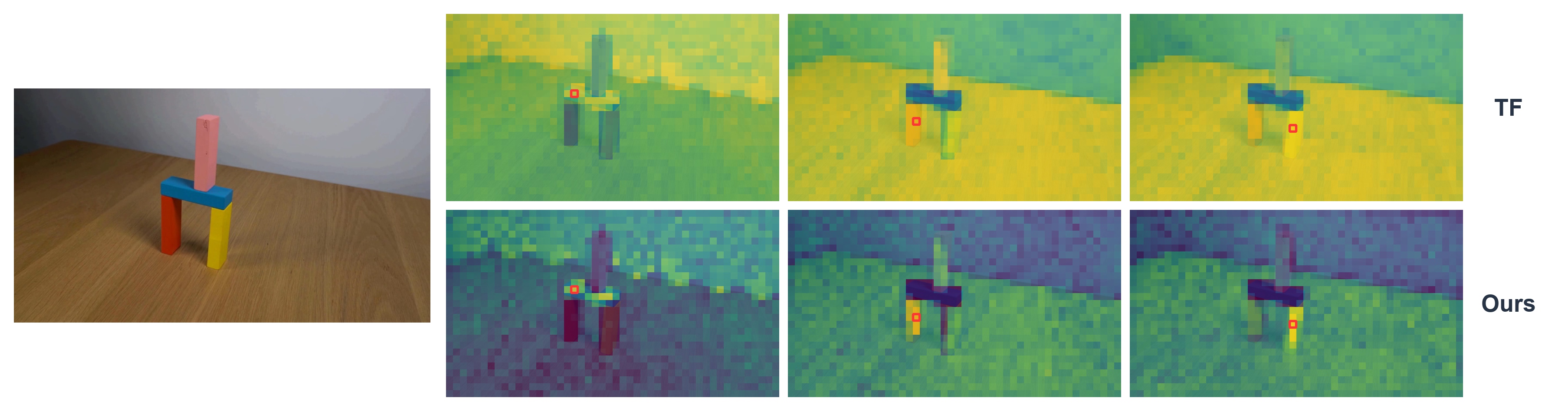}
  \caption{\textbf{Additional Spatial Feature Similarity Maps (continued).}
  Two further examples using the same query-to-token similarity visualization.}
\end{figure}

Figure~\ref{fig:representation-space-visualizations} supplements Figure~\ref{fig:qualitative-comparison} with four additional scenes.
Across these examples, FCP-trained representations tend to produce more selective similarity responses around the queried content than the teacher-forced baseline, providing further qualitative support for the spatial structure discussed in Section~\ref{sec:representation-space-analysis}.

\begin{figure}[!ht]
  \centering
  \includegraphics[width=\textwidth]{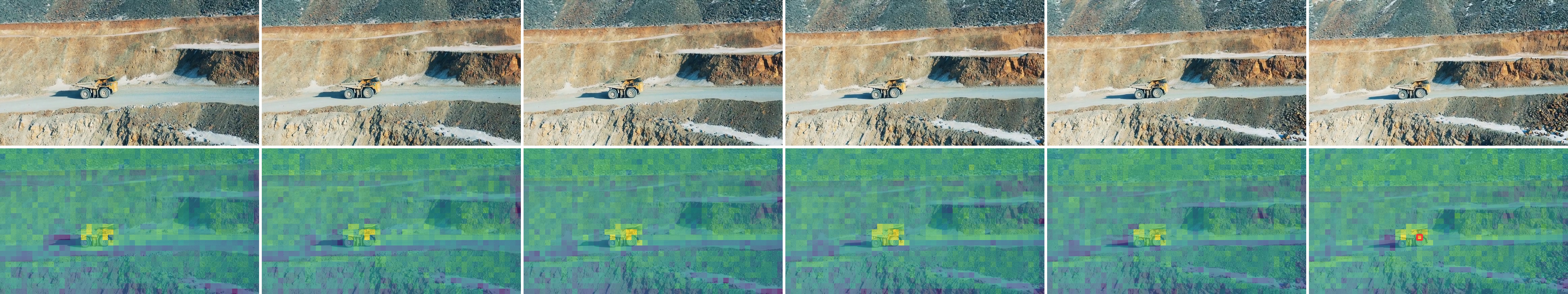}
  \par\smallskip
  \includegraphics[width=\textwidth]{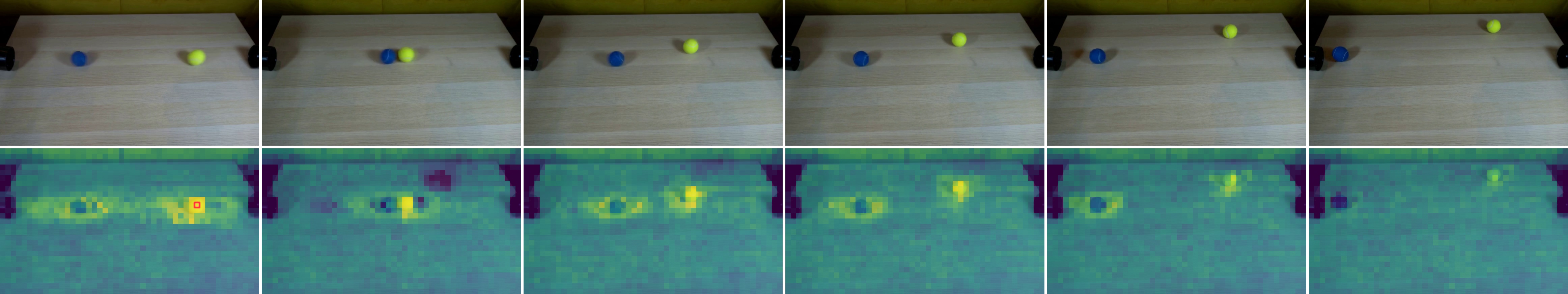}
  \par\smallskip
  \includegraphics[width=\textwidth]{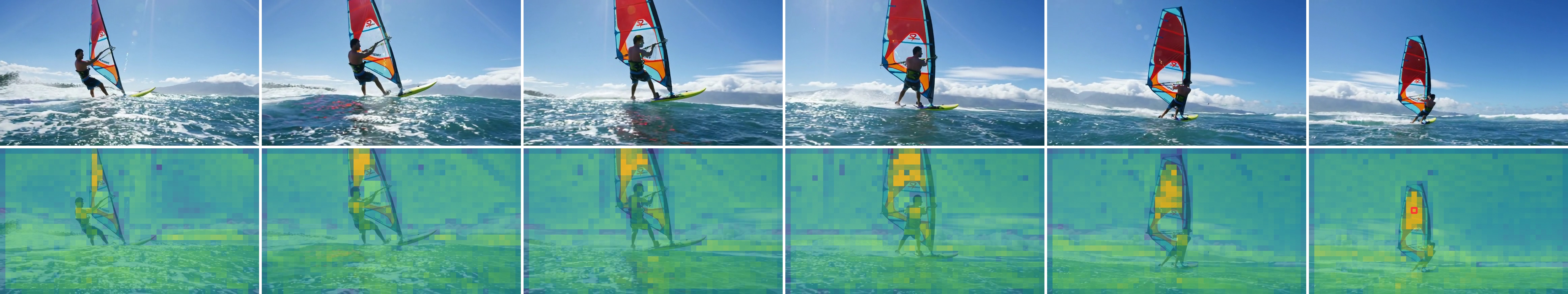}
  \caption{\textbf{Additional Spatiotemporal Feature Similarity Maps.}
  In each example, the top row shows video frames, while the bottom row visualizes similarity to a selected query token across space and time.}
  \label{fig:temporal-similarity-additional}
\end{figure}

Complementing the temporal correspondence discussed in Section~\ref{sec:representation-space-analysis}, Figure~\ref{fig:temporal-similarity-additional} shows three additional scenes in which high-similarity regions remain aligned with corresponding visual content across frames.

\begin{figure}[!ht]
  \centering
  \includegraphics[width=\textwidth]{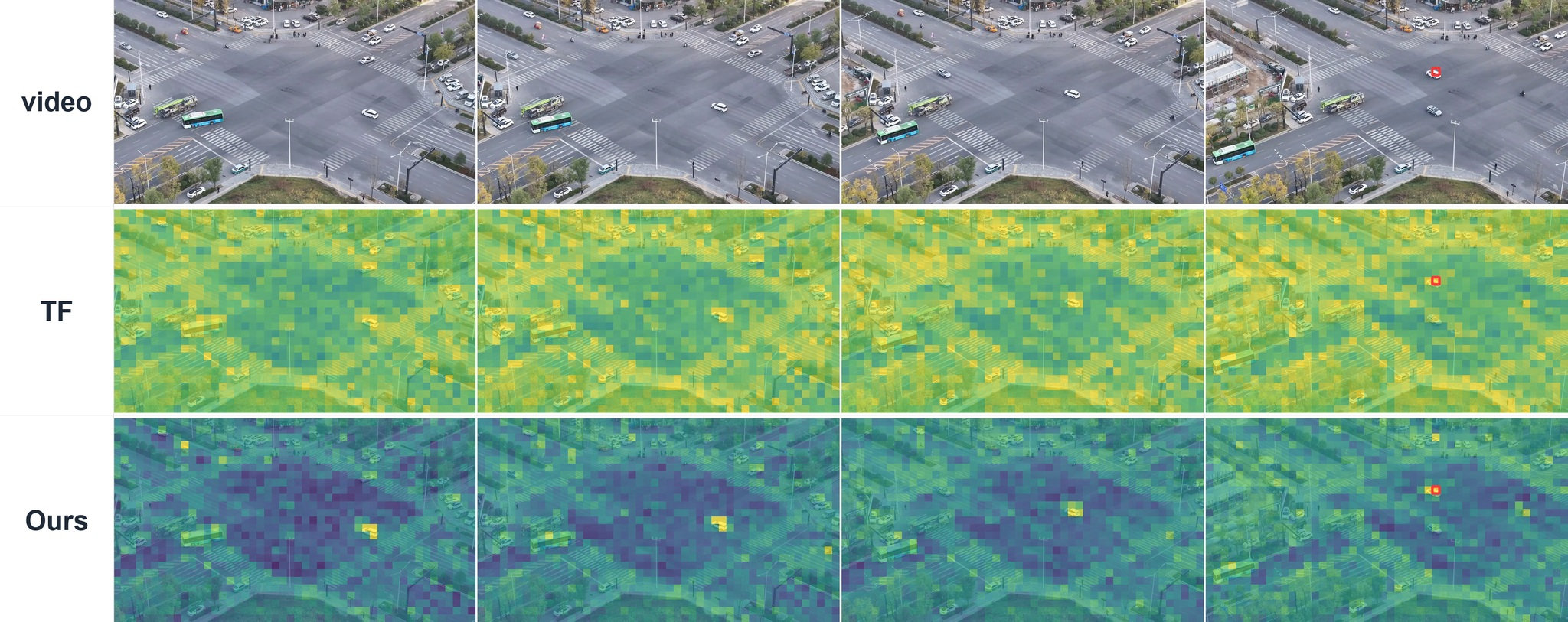}
  \par\smallskip
  \includegraphics[width=\textwidth]{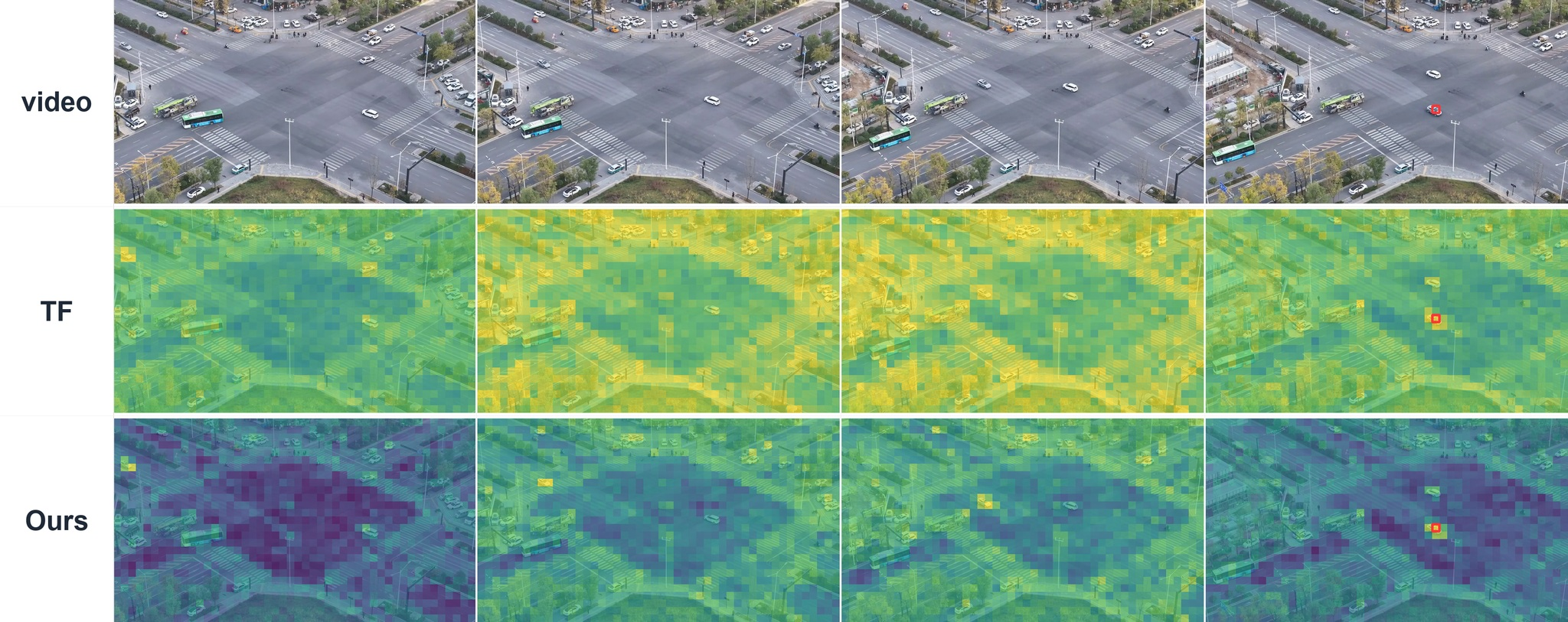}
  \caption{\textbf{Detailed Spatiotemporal Feature Similarity Maps.}
  In each example, the top row shows video frames, while the middle and bottom rows show similarity to a selected query token under teacher forcing and FCP, respectively.}
  \label{fig:detail-temporal-similarity}
\end{figure}

Figure~\ref{fig:detail-temporal-similarity} extends Figure~\ref{fig:temporal-similarity-additional} to a more challenging small-object setting.
FCP maintains localized similarity responses around the queried vehicle across frames, whereas teacher forcing produces more diffuse responses that are less able to distinguish it from the background and other objects.
This contrast suggests that FCP learns representations with stronger spatiotemporal discrimination and more meaningful semantic structure.

\subsection{Auxiliary Future Prediction Visualizations}
\label{app:auxiliary-future-prediction-visualizations}

\begin{figure}[p]
  \centering
  \includegraphics[width=\textwidth]{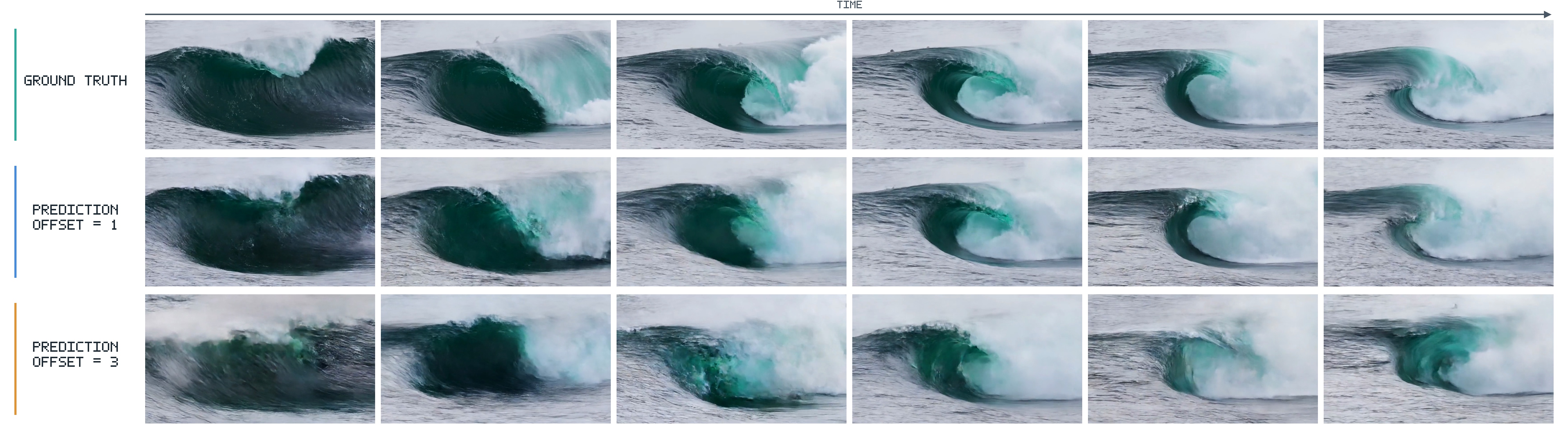}
  \par\smallskip
  \includegraphics[width=\textwidth]{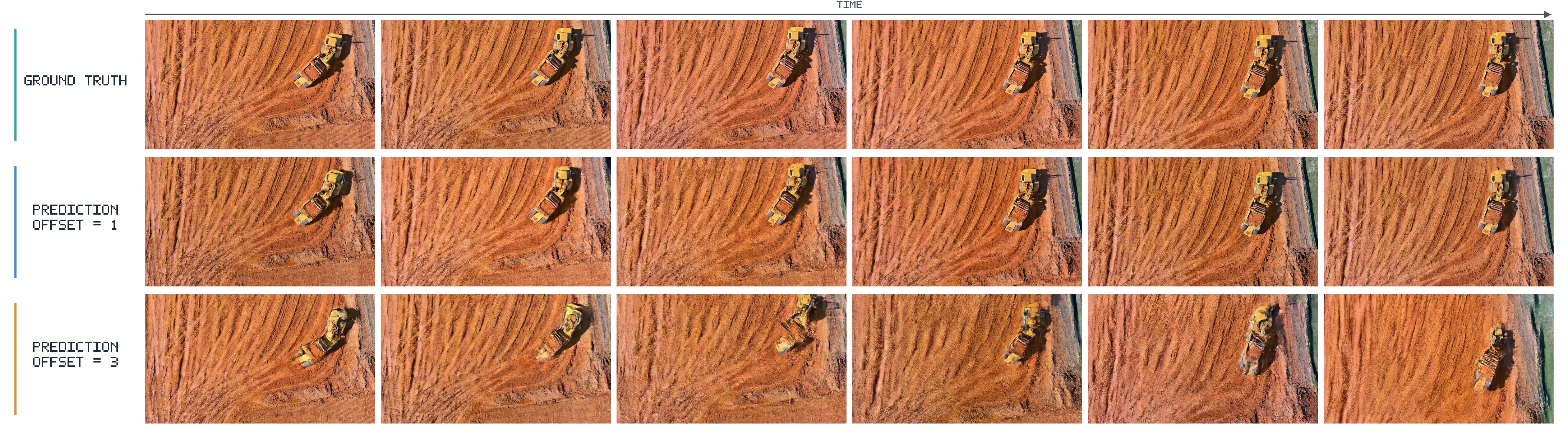}
  \par\smallskip
  \includegraphics[width=\textwidth]{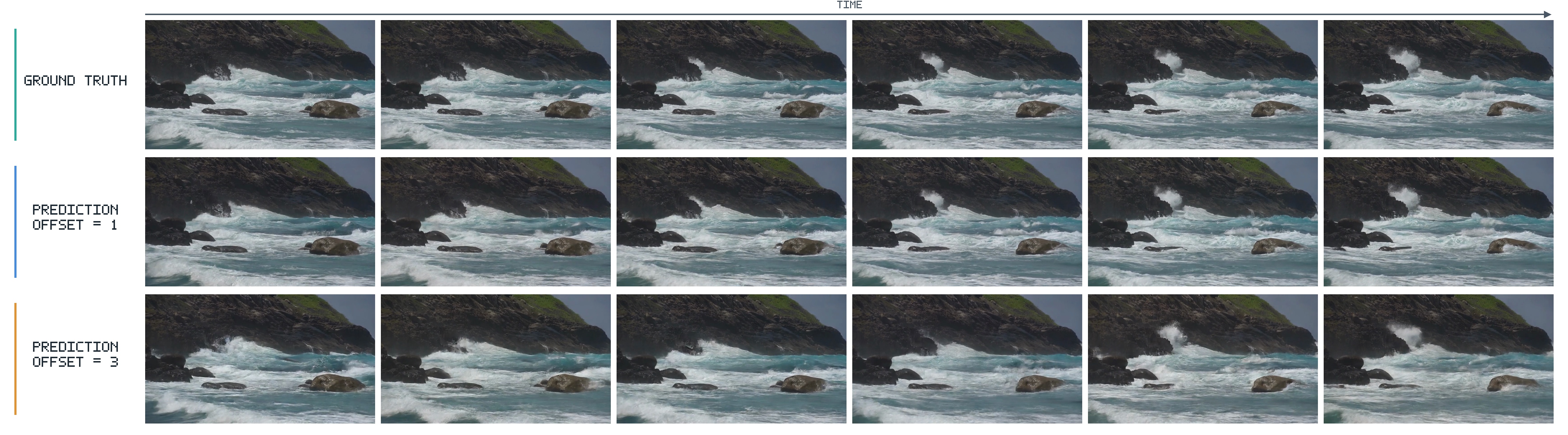}
  \par\smallskip
  \includegraphics[width=\textwidth]{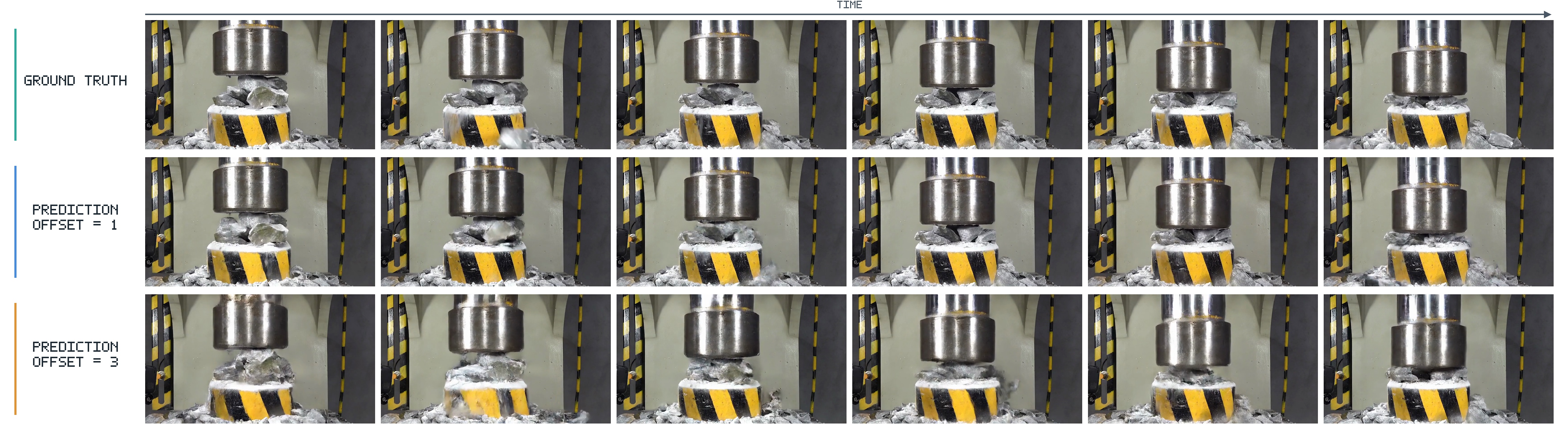}
  \par\smallskip
  \includegraphics[width=\textwidth]{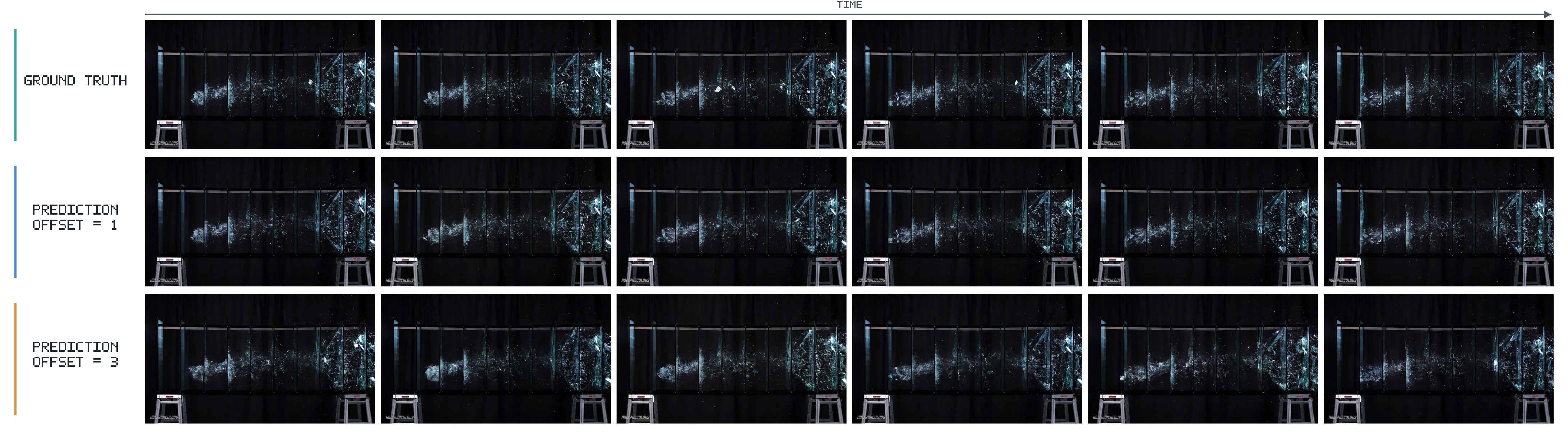}
  \caption{\textbf{Auxiliary Future Predictions.}
  In each example, the top row shows the ground-truth video, while the middle and bottom rows show outputs from auxiliary branches at the future offsets indicated on the left.}
  \label{fig:auxiliary-future-prediction-visualizations}
\end{figure}

Figure~\ref{fig:auxiliary-future-prediction-visualizations} visualizes future chunks sampled independently by the auxiliary branches from the same source representation.
Across diverse scenes, the predictions capture broad visual changes without access to the intervening chunks.
Visual fidelity tends to decline at farther offsets, yet the predictions often retain motion trends consistent with the ground truth.
These examples suggest that $\mathbf{h}_k$ contains information useful for anticipating future evolution, without implying exact reconstruction of the future frames.

\section{Limitations and Discussions}
\label{app:limitations-discussions}

\subsection{Comparison with Next Forcing}

As discussed in Section~\ref{sec:related-representation-alignment}, Next Forcing \citep{Next-Forcing} explores a closely related future-prediction idea.
Importantly, our work was developed independently of and concurrently with Next Forcing \citep{Next-Forcing}. The initial FCP prototype design was completed before the Next Forcing paper became publicly available. While both works explore future prediction for autoregressive video generation, we target different tasks and adopt distinct methodological and experimental designs. First, Next Forcing is primarily framed around world-action modeling and robotic action learning, whereas we focus on autoregressive video generation as a pathway toward generative world modeling.

\paragraph{Conditioning Context and Efficiency.}
Second, Next Forcing conditions its auxiliary predictors on the full visible history $\mathbf{h}_{\le k}$, whereas ours use only $\mathbf{h}_k$. As discussed in Section~\ref{sec:fcp}, this choice encourages the source-chunk representation itself to compactly encode information predictive of future evolution. Beyond this design rationale, we test the context choice with a full-context FCP variant shown in Figure~\ref{fig:training-convergence}. At matched training steps, the two FCP variants achieve comparable IQ scores, but ours requires roughly 40--50\% less training compute to attain the same performance. Our source-chunk-only design thus offers a more efficient and scalable form of future-prediction supervision, an important consideration when training larger video generators.

\paragraph{Parallel versus Chained Prediction.}
Third, Next Forcing adopts a chained prediction architecture inspired by language-model MTP \citep{DeepSeek-V3}, whereas our offset-specific predictors operate in parallel from $\mathbf{h}_k$. Unlike discrete-token prediction, diffusion-based video prediction gives each auxiliary predictor a partially noised future latent. Chaining may therefore pass information from an intermediate future chunk to later predictors along the red path in Figure~\ref{fig:fcp-chained}, weakening the intended predictive supervision on $\mathbf{h}_k$.

Consistent with this concern, the chained variant retains a gain at 1.3B with three offsets, but offers only a marginal I2V gain and no V2V gain at 5B models. This decline suggests that information leakage in chained prediction may become more consequential at larger model scales, as higher-capacity models can more readily exploit intermediate future information and thereby weaken the intended supervision. 
Parallel prediction, by contrast, remains effective in both settings (Tables~\ref{tab:component-ablation} and~\ref{tab:wan22-5b-experiments}), supporting its use for visual future prediction.

\paragraph{Other Differences.}
Our different task emphasis also shapes several design and evaluation choices.
We supervise both clean-context and noisy-target representations, and use token-wise AdaLN to inject source features into the auxiliary predictors, drawing on prior experience in visual generation \citep{DDT,RepFusion}.
Our evaluation focuses on the physical plausibility of generated videos, reflecting our interest in visual evolution rather than action-planning performance.

\subsection{Potential Inference-Time Applications}
\label{app:potential-inference-time-applications}

Figure~\ref{fig:auxiliary-future-prediction-visualizations} suggests that the auxiliary predictors can anticipate future dynamics, although their outputs do not yet reach the visual fidelity expected of final video generation.
Replacing the main generator with these branches throughout denoising may therefore compromise output quality.
One possibility is to use them during early, high-noise denoising before handing the sample to the main generator for refinement, potentially reducing sampling cost.
Alternatively, combining their predictions with those of the main generator may improve the reliability of the resulting future evolution.
These inference-time applications remain untested and are left for future work.

\subsection{From-Scratch Pretraining and Further Scaling}
\label{app:from-scratch-pretraining-and-scaling}

Due to computational constraints, our experiments do not cover larger model scales or training on substantially longer videos.
We therefore leave broader scaling and evaluation across longer temporal horizons to future work.

All current models are initialized from pretrained bidirectional video diffusion checkpoints rather than pretrained from scratch.
In our experiments, adapting these checkpoints for autoregressive training led to early-stage performance degradation.
We suspect that bidirectional and autoregressive models may organize their representations differently, so this initialization could limit the extent to which FCP's benefits emerge.
We aim to investigate large-scale from-scratch pretraining with FCP as a native objective for autoregressive world models, together with more diverse, higher-quality data tailored to this setting.

\end{document}